\pdfoutput=1 % for arXiv (force PDFLaTeX)
\documentclass{article}
\usepackage{wrapfig}
\PassOptionsToPackage{numbers,compress}{natbib}
\usepackage[main,final]{neurips_2026}
\usepackage{pifont}
\newcommand{\cmark}{\ding{51}}
\newcommand{\xmark}{\ding{55}}
\usepackage[utf8]{inputenc}
\usepackage[T1]{fontenc}
\usepackage{hyperref}
\usepackage{url}
\usepackage{booktabs}
\usepackage{amsfonts}
\usepackage{amsmath}
\usepackage{amssymb}
\usepackage{nicefrac}
\usepackage{microtype}
\usepackage{xcolor}
\usepackage[table]{xcolor}
\usepackage{multirow}
\usepackage{makecell}
\usepackage{array}
\usepackage{graphicx}
\usepackage{tabularx}
\usepackage{float}
\newsavebox{\uasctabbox}
\newsavebox{\fusiontabbox}

\usepackage{multirow}
\newlength{\abltabwidth}
\usepackage{tikz}
\usepackage{caption}
\usetikzlibrary{calc}
\title{RETR: A Structure-Preserving RGB-Event Transformer for Robust 3D Lane Detection}
\author{%
  Jingtao Dong, Hao Zhuang, Hao Yang,
  Liyuan Pan\thanks{Corresponding author.} , Wei Liang\\
  \normalfont Beijing Institute of Technology, Beijing, China\\
  \texttt{\{jingtao.dong, hao.zhuang\}@bit.edu.cn}\\
  \texttt{\{hao.yang, liyuan.pan, liangwei\}@bit.edu.cn}%
}
\begin{document}
\setcounter{footnote}{1}
\maketitle
\setcounter{footnote}{0}

\begin{abstract}
Robust 3D lane detection requires accurate metric road geometry recovery from visual inputs, yet conventional RGB-based approaches suffer from fragile lane evidence under adverse illumination, motion blur, and long-range perspective compression. Event cameras offer complementary high-temporal-resolution, high-dynamic-range structural cues that are robust to these challenges, but their sparse, motion-dependent responses cannot be directly fused into consistent 3D geometry. In this paper, we present the first attempt to introduce event cameras into 3D lane detection. To enable systematic research, we first build two multimodal benchmarks with metric 3D lane annotations: DSEC-3DLD (real-world sequences) and Ev-OpenLane (large-scale simulated sequences). We further propose RETR, a structure preserving transformer that converts complementary RGB and event observations into coherent 3D lane geometry. RETR first aligns modality consistent evidence through Reciprocal Context Flow Fusion, then preserves thin and uncertain lane structures with Uncertainty Aware Structural Consolidation, and finally decodes ordered lane hypotheses using a Geometry State Decoder with proposal conditioned initialization, reference conditioned query evolution, and diversified geometric inquiry. Extensive experiments show that RETR achieves state-of-the-art performance on both benchmarks, especially showing strong improvements in challenging lighting and complex road geometry scenarios. Code and dataset will be available at \url{https://github.com/duskswell/RETR}.
% Reliable 3D lane detection requires recovering metric road geometry from visual evidence that is often fragile in real driving scenes. Existing methods mainly rely on RGB frames, where lane markings can become weak or structurally incomplete under adverse illumination, motion blur, and long range perspective compression. In this paper, we make the first attempt to introduce event cameras into 3D lane detection. To enable systematic study, we establish two multimodal benchmarks with image and event inputs: DSEC-3DLD, built from real event camera driving sequences, and Ev-OpenLane, a large scale benchmark with simulated event observations. Both provide synchronized inputs and metric 3D lane annotations. 
% , reaching 75.6\% F1 score on DSEC-3DLD and 66.7\% F1 score on Ev-OpenLane. 
% These results validate the effectiveness of structure preserving RGB and event reasoning and demonstrate the importance of event cameras for robust 3D lane detection.
\end{abstract}

% \begin{figure*}[t]
%     \centering
%     \includegraphics[width=\textwidth]{pic/datasetfinal2.pdf}
%     \vspace{-2mm}
%     \caption{
%     \textbf{Overview of our RGB-event 3D lane detection dataset.}
%     Each column presents a synchronized driving sample, and each row shows, from top to bottom, the RGB image, event frame, 2D lane annotation, and metric 3D lane annotation.
%     The selected examples cover diverse real-world conditions, including tunnel scenes, low illumination, urban roads, curved lanes, and complex road layouts.
%     The event frames provide complementary high-temporal-resolution structural cues, especially when RGB images suffer from saturation, low contrast, or challenging illumination.
%     Together with image-plane 2D annotations and metric 3D lane labels, the dataset enables systematic evaluation of RGB-event fusion for geometry-aware 3D lane detection.
%     }
%     \label{fig:dataset_overview}
%     \vspace{-2mm}
% \end{figure*}
\section{Introduction}

% Reliable 3D lane detection is a fundamental perception task for autonomous driving. Unlike 2D
% lane detection, which localizes lane markings in the image plane, 3D lane detection must recover
% metric lane geometry in the physical world, including lateral position, longitudinal layout,
% elevation variation, visibility, and lane continuity. 
Reliable 3D lane detection is a core perception task
for autonomous driving. It aims to recover metric
lane geometry, including lateral position, longitudinal
layout, and elevation, together with lane visibility
and continuity. Recent methods have advanced this task
through stronger geometric representations
~\cite{guo2020gen}, camera-to-ground transformations
and BEV construction~\cite{chen2022persformer,yao2023sparse},
anchor-based lane modeling~\cite{huang2023anchor3dlane},
and query-based decoding with transformer-based
long-range reasoning~\cite{luo2023latr,luo2024dv}.
However, their predictions still depend largely on
the lane evidence available in RGB frames.
This evidence is often fragile in real driving scenes:
lane markings are thin, low-contrast, and strongly
compressed by perspective, especially at long range.
Under adverse illumination and motion conditions,
lane evidence can become weak or structurally incomplete,
forcing the model to recover 3D geometry from
unreliable visual cues.

The broader challenge of perception from degraded
observations has motivated research on adverse-weather
removal~\cite{yang2024language}, defocus
deblurring~\cite{yang2024ldp}, and fog-invariant action
recognition~\cite{Liu_2026_CVPR}.
For metric 3D lane detection, however, improving visual
appearance must also preserve the geometric evidence
needed for localization. A recent study of GPT-Image
for image restoration~\cite{yang2025preliminary}
reports that perceptually plausible outputs can deviate
from the original image geometry and perspective.
This distinction motivates our focus on acquiring
complementary structural observations and preserving
their geometric information for 3D lane recovery.

Event cameras provide a promising complementary modality. Different from frame-based cameras that capture intensity at fixed intervals, event cameras asynchronously record local brightness changes with microsecond-level temporal resolution and extremely high dynamic range. They produce boundary-aware responses that remain stable under fast motion, over-exposure, under-exposure, and low-contrast conditions \cite{chen2020event,li2019event,maqueda2018event}.
% Event cameras offer a complementary way to observe such fragile structures. By asynchronously
% recording local brightness changes with high temporal resolution and high dynamic range, they can
% retain boundary like responses under fast motion and difficult illumination \cite{chen2020event,li2019event,maqueda2018event}. 
These responses are
particularly relevant to lane perception, where distant markings, road boundaries, and weak contrast
transitions may disappear in conventional frames. 
\begin{wrapfigure}{r}{0.55\textwidth}

\centering
\includegraphics[width=\linewidth]{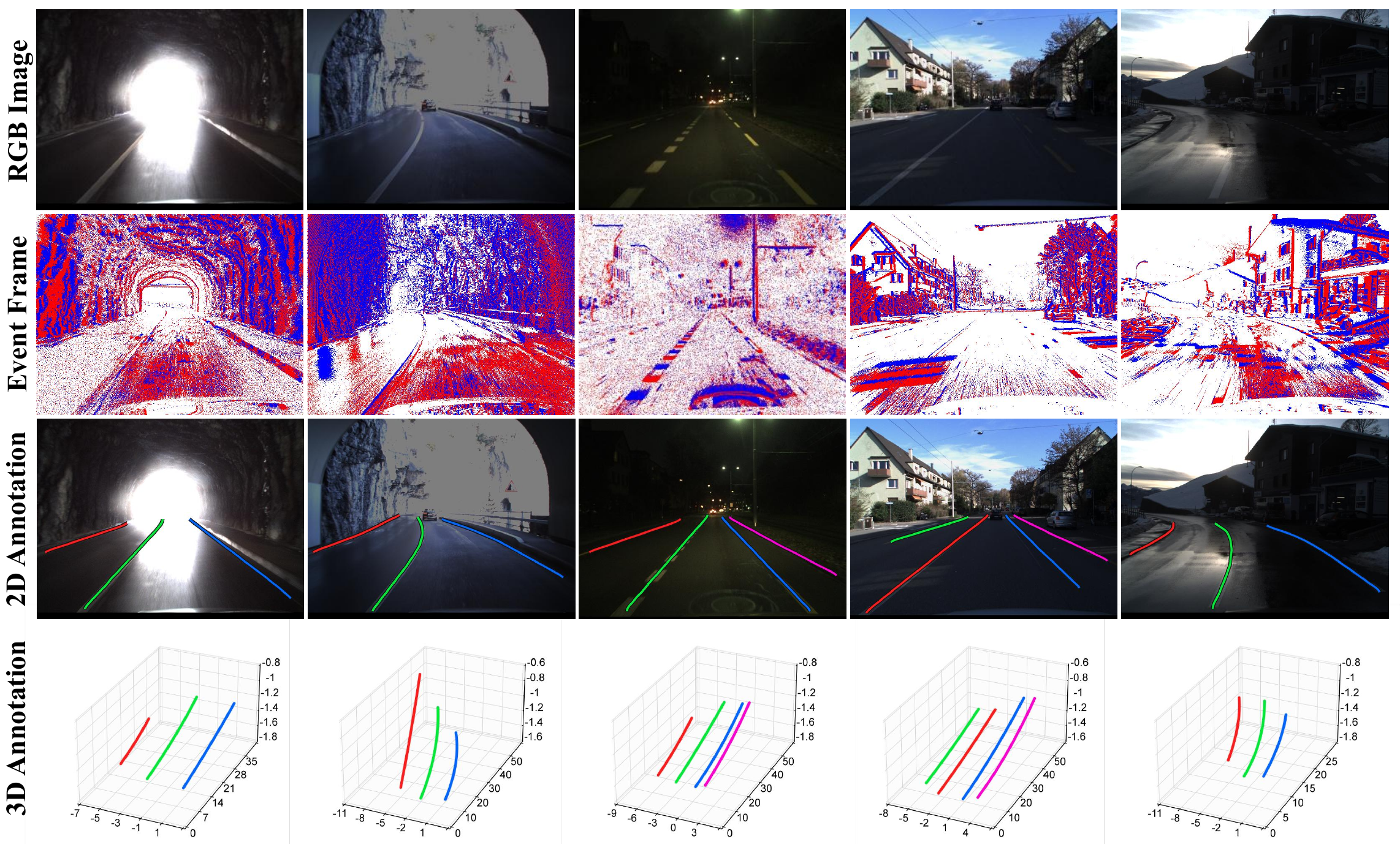}

\caption{
Overview of our DSEC-3DLD dataset.
Each column presents one synchronized driving sample, including the RGB image, event frame,
2D lane annotation, and metric 3D lane annotation.
The selected examples cover challenging real-world conditions such as tunnels, low illumination and strong exposure changes.
}
\label{fig:dataset_overview}

\end{wrapfigure}
% Yet event signals do not directly translate into
% reliable 3D lane geometry. They are sparse, polarity-driven, motion-dependent, and substantially
% different from RGB appearance features 
Yet event signals are sparse, motion-dependent and structurally different from RGB features \cite{gallego2020event,chakravarthi2024recent}. Therefore, a simple fusion strategy, such as feature
concatenation, can easily dilute useful event responses or amplify unreliable activations, leading
to geometrically inconsistent lane predictions.

These observations motivate a structure preserving view of 3D lane detection with RGB and event
inputs. The goal is not merely to introduce an additional event branch, but to transform
complementary sensing evidence into coherent metric lane geometry. This requires a full pipeline
that aligns modality specific cues, preserves weak but important lane structures during feature
aggregation, and decodes the resulting representation as ordered 3D lane hypotheses. To this end, we propose RETR, a structure preserving transformer for robust 3D lane detection.
Rather than treating RGB and event inputs as features to be simply merged, our approach organizes
the prediction process around a progressive transformation from complementary sensing evidence to
metric lane geometry. It promotes modality consistent responses and bidirectional context exchange
across RGB and event features through Reciprocal Context Flow Fusion (RCF), while Uncertainty Aware
Structural Consolidation (UASC) retains thin, uncertain, and event induced lane evidence during
multi scale aggregation. The consolidated representation is further decoded by the Geometry State
Decoder (GSD), which models lanes as ordered geometric entities using proposal conditioned
initialization, reference conditioned query evolution, and diversified geometric inquiry. In this
way, our approach turns complementary RGB and event observations into stable 3D lane geometry.

A key obstacle to studying this setting is the lack of dedicated benchmarks. Existing 3D lane
datasets are primarily built with conventional frame cameras, making them insufficient for evaluating
the role of event cameras in 3D lane detection. To the best of our knowledge, we are the first to
introduce event cameras into 3D lane detection. We further establish two multimodal benchmarks with
image and event inputs: DSEC-3DLD, a real event camera dataset with synchronized RGB and event inputs
and metric 3D lane annotations, with representative examples shown in
Fig.~\ref{fig:dataset_overview}, and Ev-OpenLane, a large scale benchmark with simulated event
information for scalable training and evaluation. Experiments on both benchmarks show that RETR
achieves state-of-the-art performance, validating the effectiveness of our structure preserving
design and demonstrating the importance of event cameras for robust 3D lane detection.

Our contributions are summarized as follows:
\begin{itemize}
    \item We introduce event-based 3D lane detection and establish two multimodal benchmarks, DSEC-3DLD and Ev-OpenLane, providing synchronized image and event inputs together with metric 3D lane annotations. To the best of our knowledge, this is the first work to study event cameras for 3D lane detection.

    \item We propose RETR, a structure preserving transformer that transforms complementary RGB and
    event observations into coherent 3D lane geometry.

    \item We develop a reliability-to-geometry pipeline that integrates RCF, UASC and GSD to align
cross-modal lane evidence, preserve sparse structural cues, and decode coherent 3D lanes through
proposal guided geometry state evolution.
\end{itemize}

\section{Related Work}

\textbf{3D lane detection.}
% Lane detection has evolved from 2D image-space formulations to geometry-aware 3D reasoning. In 2D lane detection, existing methods can be broadly grouped into segmentation-based approaches \cite{neven2018towards, zheng2021resa, pan2018spatial}, anchor- or offset-based regression methods \cite{tabelini2021keep, zheng2022clrnet, li2019line}, keypoint-based representations \cite{ko2021key, qu2021focus, wang2022keypoint, jin2022eigenlanes}, and curve-parameterized formulations \cite{van2019end, tabelini2021polylanenet, liu2021end, feng2022rethinking}. While these methods are effective for image-plane localization, they do not explicitly recover the 3D structure required by autonomous driving.
Advances in deep learning have substantially expanded
visual understanding capabilities~\cite{liu2026storyboard, han2025dynfocus, han2026dycpo,
han2026vedas, liu2026benchmarking}.
In 3D lane detection, recent methods have improved
geometric reasoning through explicit spatial
transformation or implicit structure modeling. Explicit approaches project image features into bird's-eye-view space using inverse perspective mapping or depth-aware lifting~\cite{chen2022persformer, yao2023sparse, guo2020gen, li2022reconstruct, pittner20233d, pittner2024lanecpp, ghafoorian2018gan, mallot1991inverse, luo2022m}, sometimes further incorporating depth cues, geometric constraints, height-aware representations, or slope modeling~\cite{yan2022once, guo2020gen, park2025heightlane, park2025sc}. In contrast, implicit methods directly infer 3D lane structures with queries~\cite{liu2022petr, luo2023latr, luo2024dv}, 3D anchors~\cite{huang2023anchor3dlane}, or BEV keypoint grouping~\cite{wang2023bev, ozturk2025glane3d}. Despite these advances, existing methods still derive their lane evidence primarily from frame-based intensity images, making them vulnerable to severe illumination degradation such as overexposure, nighttime underexposure, headlight glare, and abrupt lighting transitions.
% Recent 3D lane detection methods address this limitation through either explicit spatial transformation or implicit geometric reasoning. Explicit methods project image features into a bird's-eye-view representation via inverse perspective mapping or depth-aware lifting \cite{chen2022persformer, yao2023sparse, guo2020gen, li2022reconstruct, pittner20233d, pittner2024lanecpp, ghafoorian2018gan, mallot1991inverse, luo2022m}, while other approaches combine image predictions with depth or geometric constraints \cite{yan2022once, guo2020gen}. To better model elevation changes and complex road topology, several methods further incorporate height-aware representations or slope modeling \cite{park2025heightlane, park2025sc}. In parallel, implicit paradigms avoid explicit warping and directly infer 3D lane structure using queries \cite{liu2022petr, luo2023latr, luo2024dv}, 3D anchors \cite{huang2023anchor3dlane}, or keypoint grouping in BEV space \cite{wang2023bev, ozturk2025glane3d}. Despite their strong geometric modeling ability, these methods remain largely reliant on RGB imagery, making them inherently susceptible to extreme lighting conditions, including severe overexposure, nighttime underexposure, headlight glare, and sudden illumination transitions.

\textbf{Event-based vision for robust perception.}
Event cameras asynchronously capture intensity changes
with microsecond latency and high dynamic range, making
them attractive for perception in challenging conditions.
Their applications, datasets, and practical challenges
in automotive sensing are reviewed by Shariff
et al.~\cite{shariff2024event}.
They have been successfully applied to object detection
\cite{mitrokhin2018event, zhou2023rgb, zhou2024event,
lin2024spike}, segmentation
\cite{alonso2019ev, mitrokhin2018event,
zhu2024continuous, stoffregen2019event}, optical flow
\cite{pan2020single, ding2022spatio, luo2024efficient,
zhu2018ev}, and image restoration
\cite{pan2019bringing, sun2022event, jiang2020learning,
zhou2023deblurring}.
In autonomous driving, event-based datasets such as
DSEC~\cite{gehrig2021dsec} and
MVSEC~\cite{zhu2018multivehicle} provide synchronized
event, frame, and geometric measurements for downstream
multimodal perception.
NSAVP~\cite{carmichael2025nsavp} further combines event,
thermal, monochrome, and RGB observations and benchmarks
place recognition under varied lighting conditions.
More recently, event signals have also been explored
for 3D object detection
\cite{cho2025ev, kang2025unleashing}.
For lane perception, DET~\cite{cheng2019det} provides
a high-resolution event-based lane extraction benchmark
with pixel-level annotations. While DET addresses
image-plane lane extraction, our work targets metric
3D lane geometry from RGB-event observations.

\textbf{Lane datasets.}
Benchmark development has similarly progressed from 2D lane datasets such as TuSimple \cite{tusimple2017}, CULane \cite{pan2018spatial}, and LLAMAS \cite{behrendt2019unsupervised} to 3D benchmarks such as Apollo-Sim-3D \cite{guo2020gen}, ONCE-3DLanes \cite{yan2022once}, and OpenLane  \cite{chen2022persformer}. These datasets have greatly advanced geometric lane perception, but they are primarily built on conventional frame cameras. As a result, they provide limited support for studying multimodal robustness under adverse conditions. Our work is motivated by this gap: we study RGB-event fusion for 3D lane detection and build a benchmark setting tailored to robust 3D road topology understanding in challenging environments.

\section{Method}

\paragraph{Overview.}
Our approach builds on LATR~\cite{luo2023latr} with a dual-stream design for RGB and event inputs.
Given a synchronized RGB image and an event representation aggregated within a short temporal window,
two modality-specific encoders extract hierarchical features
$\{\mathbf{F}_{r}^{l}\}_{l=1}^{L}$ and $\{\mathbf{F}_{e}^{l}\}_{l=1}^{L}$, where $l$ indexes the
feature stage and $L$ is the number of selected stages. RETR first aligns and exchanges RGB and
event features through RCF. The resulting multi-scale features are then consolidated by UASC to
preserve weak lane structures before decoding. Finally, GSD transforms the consolidated
representation into ordered 3D lane hypotheses, including lane existence, anchor-wise geometry, and
visibility.

\subsection{Reciprocal Context Flow Fusion}

Let $\mathbf{F}_{r}^{l},\mathbf{F}_{e}^{l}\in\mathbb{R}^{B\times C_l\times H_l\times W_l}$ denote
the RGB and event features at stage $l$, where $B$ is the batch size, $C_l$ is the channel dimension,
and $H_l,W_l$ are the spatial resolution. RGB features encode dense appearance and semantic context,
while event features capture temporal changes and high-frequency structures. Because these modalities
respond differently to lane evidence, naive concatenation can mix redundant or misaligned responses.
RCF therefore starts by mapping both modalities into a shared stage-wise feature space before
co-activation enhancement and bidirectional context exchange:
\begin{equation}
\bar{\mathbf{F}}_{r}^{l}
=
\psi_{r}^{l}(\mathbf{F}_{r}^{l}),
\qquad
\bar{\mathbf{F}}_{e}^{l}
=
\psi_{e}^{l}(\mathbf{F}_{e}^{l}),
\end{equation}
where $\psi_{r}^{l}$ and $\psi_{e}^{l}$ are $1\times1$ convolutional projections. A co-activation
response is then computed by element-wise multiplication:
\begin{equation}
\mathbf{M}^{l}
=
\bar{\mathbf{F}}_{r}^{l}
\odot
\bar{\mathbf{F}}_{e}^{l}.
\end{equation}
The response $\mathbf{M}^{l}$ emphasizes spatial locations that are supported by both RGB appearance
and event structures. Rather than using it as a hard mask, we inject it into both streams by residual
addition:
\begin{equation}
\mathbf{H}_{r}^{l}
=
\bar{\mathbf{F}}_{r}^{l}
+
\mathbf{M}^{l},
\qquad
\mathbf{H}_{e}^{l}
=
\bar{\mathbf{F}}_{e}^{l}
+
\mathbf{M}^{l}.
\end{equation}
This produces co-activated RGB and event features while keeping the two modalities distinguishable.

To exchange complementary information, RCF uses a Cross-Attention Block (CAB). Given a target feature
$\mathbf{X}$ and a guidance feature $\mathbf{Y}$, CAB is implemented as
\begin{equation}
\operatorname{CAB}(\mathbf{X}\mid\mathbf{Y})
=
\mathbf{X}
+
\rho
\left(
\left[
\mathbf{X},
\operatorname{Unflat}
\left(
\operatorname{softmax}
\left(
\frac{(W_q\mathbf{X})(W_k\mathbf{Y})^{\top}}{\sqrt{d}}
\right)
W_v\mathbf{Y}
\right)
\right]
\right).
\end{equation}
Before attention, $\mathbf{X}$ and $\mathbf{Y}$ are flattened over spatial positions. The target
feature provides the query through $W_q$, while the guidance feature provides keys and values through
$W_k$ and $W_v$. The operator $\operatorname{Unflat}(\cdot)$ restores the spatial layout after
attention, $d$ is the key dimension, and $\rho(\cdot)$ is a Conv-BN-ReLU fusion block. CAB therefore
transports context from the guidance modality and fuses it back into the target modality.

RCF applies CAB in both directions, generating an event-guided RGB feature
$\mathbf{Y}_{r}^{l}=\operatorname{CAB}(\mathbf{H}_{r}^{l}\mid\mathbf{H}_{e}^{l})$ and an
RGB-guided event feature
$\mathbf{Y}_{e}^{l}=\operatorname{CAB}(\mathbf{H}_{e}^{l}\mid\mathbf{H}_{r}^{l})$. The two features
are concatenated along the channel dimension and compressed by a $1\times1$ Conv-BN-ReLU projection
to obtain the fused stage feature $\mathbf{Z}^{l}$. The fused hierarchy
$\{\mathbf{Z}^{l}\}_{l=1}^{L}$ is then passed to UASC.

\begin{figure*}[t]
    \centering
    \begin{tikzpicture}

        \node[anchor=south west, inner sep=0] (methodimg) at (0,0)
        {\includegraphics[width=\textwidth]{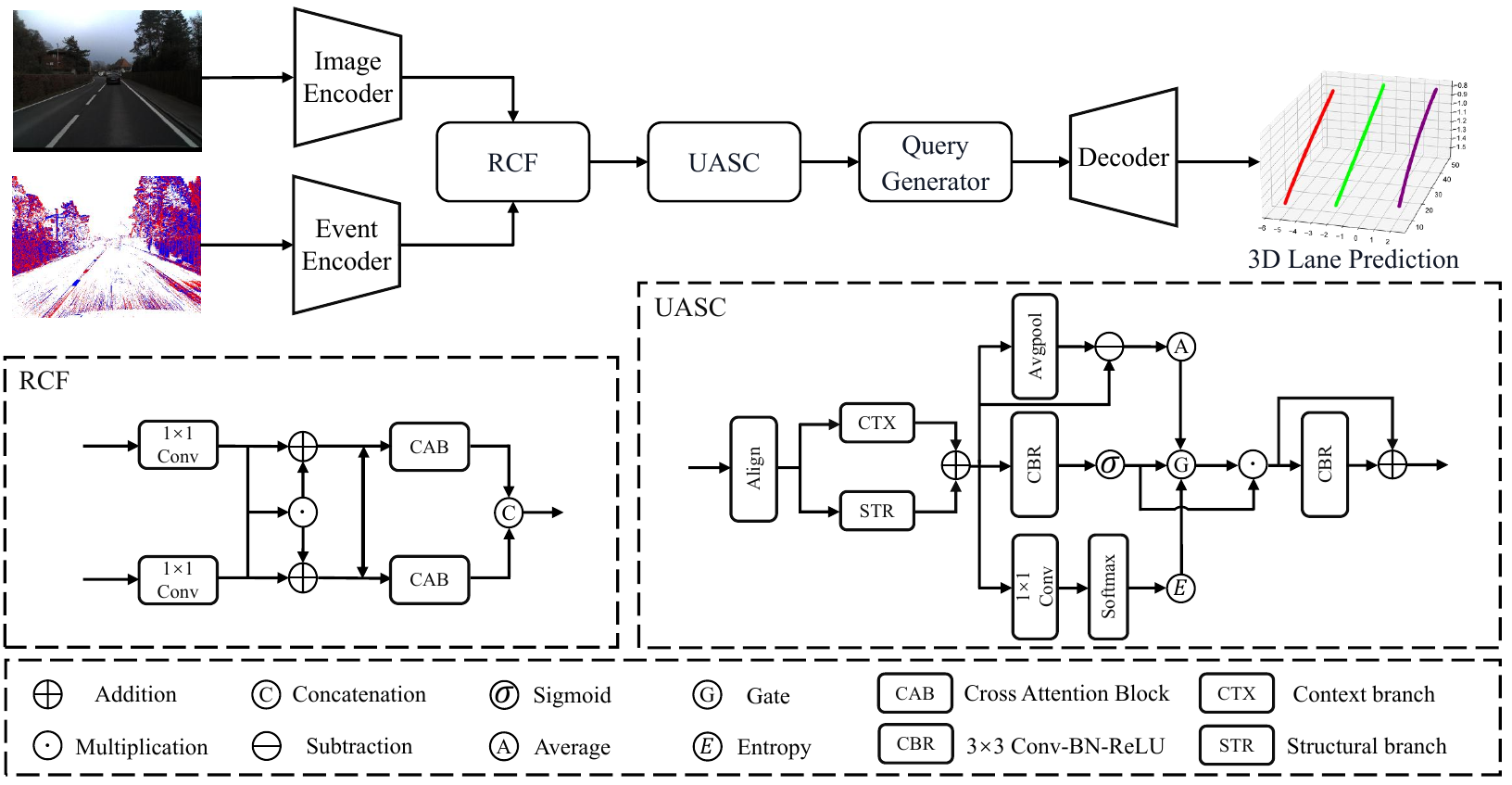}};

        \begin{scope}[x={(methodimg.south east)}, y={(methodimg.north west)}]

            \tikzset{
                methodvar/.style={
                    font=\scriptsize,
                    inner sep=0.8pt
                },
                methodvarsmall/.style={
                    font=\tiny,
                    inner sep=0.5pt
                }
            }

            % ---------- Top pipeline ----------
            \node[methodvar] at (0.07,0.583)
            {Event};
            \node[methodvar] at (0.07,0.788)
            {Image};
            \node[methodvar] at (0.305,0.935)
            {$\{\mathbf{F}_{r}^{l}\}_{l=1}^{L}$};
            \node[methodvar] at (0.545,0.405)
            {$\mathbf{Z}_a$};
            \node[methodvar] at (0.305,0.655)
            {$\{\mathbf{F}_{e}^{l}\}_{l=1}^{L}$};

            \node[
                font=\scriptsize,
                inner sep=0.5pt
            ] at (0.412,0.721)
            {$\{\mathbf{Z}^{l}\}_{l=1}^{L}$};

            \node[methodvar] at (0.55,0.755)
            {$\mathbf{F}_{c}$};

            % ---------- RCF ----------
            \node[methodvar] at (0.07,0.458)
            {$\mathbf{F}_{r}^{l}$};

            \node[methodvar] at (0.07,0.288)
            {$\mathbf{F}_{e}^{l}$};

            \node[methodvar] at (0.225,0.458)
            {$\mathbf{H}_{r}^{l}$};
            \node[methodvar] at (0.225,0.354)
            {$\mathbf{M}^{l}$};
            \node[methodvar] at (0.225,0.235)
            {$\mathbf{H}_{e}^{l}$};
            \node[methodvar] at (0.357,0.375)
            {$\mathbf{Z}^{l}$};

            \node[methodvar] at (0.69,0.721)
            {\scalebox{0.9}{$\{\mathbf{Q}_{i}^{0},\mathbf{r}_{i}^{0}\}_{i=1}^{N}$}};

            % ---------- UASC ----------
            \node[
                fill=none,
                draw=none,
                inner sep=0pt
            ] at (0.456,0.366)
            {\scalebox{0.65}{$\{\mathbf{Z}^{l}\}_{l=1}^{L}$}};

            \node[methodvarsmall] at (0.8,0.58)
            {$\mathbf{A}$};

            \node[methodvarsmall] at (0.753,0.422)
            {$\mathbf{D}$};
            \node[methodvarsmall] at (0.659,0.386)
            {\scalebox{0.9}{$\mathbf{F}_b$}};
            \node[methodvarsmall] at (0.8,0.422)
            {$\boldsymbol{\Omega}$};
            \node[methodvarsmall] at (0.8,0.225)
            {$\mathbf{E}$};

            \node[methodvar] at (0.942,0.384)
            {$\mathbf{F}_{c}$};

        \end{scope}
    \end{tikzpicture}

    \caption{
    \textbf{Overall architecture of RETR.}
    RETR converts synchronized RGB and event observations into metric 3D lane geometry through a
    structure-preserving RGB-event pipeline. Modality-specific encoders first extract multi-scale RGB
    and event features $\{\mathbf{F}_{r}^{l}\}_{l=1}^{L}$ and
    $\{\mathbf{F}_{e}^{l}\}_{l=1}^{L}$. RCF aligns both modalities at each scale by using a
    co-activation response $\mathbf{M}^{l}$ to emphasize jointly supported lane evidence and
    bidirectional Cross Attention Blocks (CAB) to exchange complementary context, producing fused features
    $\{\mathbf{Z}^{l}\}_{l=1}^{L}$. UASC aggregates the fused hierarchy into $\mathbf{F}_{c}$ while
    preserving thin and uncertain lane structures. It combines context and structural branches and
    gates detail refinement using high-frequency response $\mathbf{A}$, learned detail cue
    $\mathbf{D}$, and entropy-based uncertainty $\mathbf{E}$. The query generator and geometry
    decoder finally predict ordered 3D lane hypotheses, including lane existence, 3D coordinates,
    and visibility.
    }
    \label{fig:method}
    % \vspace{-2mm}
\end{figure*}

\subsection{Uncertainty-Aware Structural Consolidation}

Given the fused hierarchy $\{\mathbf{Z}^{l}\}_{l=1}^{L}$, UASC produces a compact representation
while preserving thin and uncertain lane evidence. Each fused stage is first projected to a common
channel dimension and resized to a shared spatial resolution:
\begin{equation}
\hat{\mathbf{Z}}^{l}
=
\operatorname{Align}
\left(
\phi_l(\mathbf{Z}^{l})
\right),
\qquad
l=1,\ldots,L.
\end{equation}
The projection $\phi_l$ is implemented by a $1\times1$ convolution, and
$\operatorname{Align}(\cdot)$ denotes bilinear resizing. The aligned features are concatenated along
the channel dimension to form the aligned multi-scale feature $\mathbf{Z}_{a}$.

UASC contains two branches with different responsibilities. The structural branch, denoted as STR,
preserves lane geometry from both all-scale evidence and high-resolution details. It first fuses all
aligned stages to obtain a scale-complete representation and also processes the high-resolution
stages to retain local lane structures:
\begin{equation}
\mathbf{F}_{\mathrm{STR}}
=
\mathcal{B}_{\mathrm{STR}}
\left(
\left[
\mathcal{B}_{\mathrm{ms}}(\mathbf{Z}_{a}),
\mathcal{B}_{\mathrm{hr}}
\left(
[
\hat{\mathbf{Z}}^{1},\ldots,\hat{\mathbf{Z}}^{L_h}
]
\right)
\right]
\right).
\end{equation}
$\mathcal{B}_{\mathrm{ms}}$ fuses all aligned scales using Conv-BN-ReLU layers, while
$\mathcal{B}_{\mathrm{hr}}$ processes the first $L_h$ high-resolution stages. The output
$\mathbf{F}_{\mathrm{STR}}$ therefore keeps global scale coverage and fine spatial details, which is
important for thin lane markings and sparse event responses.

The context branch, denoted as CTX, complements STR by modeling road-level layout and long-range
lane continuity. It uses parallel dilated convolutions with different receptive fields:
\begin{equation}
\mathbf{F}_{\mathrm{CTX}}
=
\mathcal{B}_{\mathrm{CTX}}
\left(
\left[
\mathcal{D}_{1}(\mathbf{Z}_{a}),
\mathcal{D}_{2}(\mathbf{Z}_{a}),
\mathcal{D}_{5}(\mathbf{Z}_{a}),
\mathcal{D}_{9}(\mathbf{Z}_{a})
\right]
\right).
\end{equation}
Each $\mathcal{D}_{s}$ is a $3\times3$ dilated Conv-BN-ReLU block with dilation rate $s$, and
$\mathcal{B}_{\mathrm{CTX}}$ integrates the multi-range responses. By aggregating context at
multiple dilation scales, CTX provides layout-level continuity cues for weak or partially missing
lane evidence while keeping the spatial resolution unchanged.

The STR and CTX outputs are then coupled to form a structure-context base feature
$\mathbf{F}_{b}$. This coupling combines the detail-preserving response from STR with the
layout-aware continuity response from CTX, providing a stronger base representation for the
subsequent uncertainty-guided refinement.

To identify regions requiring structural refinement, UASC computes three cues from $\mathbf{F}_{b}$:
\begin{equation}
\begin{aligned}
\mathbf{E}
&=
\mathcal{H}
\left(
\operatorname{softmax}
\left(
h_c(\mathbf{F}_{b})
\right)
\right),\\
\mathbf{A}
&=
\operatorname{Avg}_{c}
\left(
\left|
\mathbf{F}_{b}
-
\operatorname{AvgPool}(\mathbf{F}_{b})
\right|
\right),\\
\mathbf{D}
&=
\sigma
\left(
h_d(\mathbf{F}_{b})
\right).
\end{aligned}
\end{equation}
The entropy cue $\mathbf{E}$ is computed from the channel-wise probability distribution predicted by
the classification projection $h_c$, the high-frequency cue $\mathbf{A}$ measures the
channel-averaged residual between $\mathbf{F}_{b}$ and its spatially averaged version, and the learned
detail cue $\mathbf{D}$ is predicted by the detail head $h_d$. These cues are then fused into a
refinement gate:
\begin{equation}
\boldsymbol{\Omega}
=
\sigma
\left(
h_g
\left(
[
\mathbf{E},
\mathbf{A},
\mathbf{D}
]
\right)
\right).
\end{equation}
The gating head $h_g$ is implemented with lightweight convolutional layers. The gate
$\boldsymbol{\Omega}$ indicates the spatial regions where the base feature requires additional
structural refinement.
The gated detail cue is first converted into a structural refinement feature:
\begin{equation}
\mathbf{F}_{\mathrm{ref}}
=
\mathcal{B}_{\mathrm{ref}}
\left(
[
\mathbf{F}_{b},
\boldsymbol{\Omega}
\odot
\mathbf{D}
]
\right).
\end{equation}
Here, $\mathbf{F}_{\mathrm{ref}}$  is a refinement feature predicted
from the base feature and the gated detail cue.
\begin{equation}
\mathbf{F}_{c}
=
\mathcal{B}_{o}
\left(
[
\mathbf{F}_{b},
\mathbf{F}_{b}
+
\boldsymbol{\Omega}
\odot
\mathbf{F}_{\mathrm{ref}}
]
\right).
\end{equation}
The final consolidated feature
$\mathbf{F}_{c}$ is obtained by residually injecting $\mathbf{F}_{\mathrm{ref}}$ into
$\mathbf{F}_{b}$ through $\boldsymbol{\Omega}$, followed by the output projection
$\mathcal{B}_{o}$. This residual design prevents uncertain detail cues from replacing the base
representation and only enhances regions selected by the refinement gate.

\subsection{Geometry State Decoder}

Given the consolidated feature $\mathbf{F}_{c}$, GSD transforms the fused lane evidence into
ordered 3D lane hypotheses. As shown in Fig.~\ref{fig:method}, the query generator maps
$\mathbf{F}_{c}$ to $N$ initial lane states
$\{\mathbf{Q}_{i}^{0},\mathbf{r}_{i}^{0}\}_{i=1}^{N}$ by first extracting proposal descriptors
$\{\mathbf{s}_{i}\}_{i=1}^{N}$ and then expanding each descriptor into ordered anchor tokens. For
proposal $i$, $\mathbf{Q}_{i}^{0}\in\mathbb{R}^{A\times C}$ is an ordered anchor-token sequence
associated with predefined longitudinal positions, and
$\mathbf{r}_{i}^{0}\in[0,1]^{A\times2}$ provides the initial lateral-height reference for these
anchors. The decoder then evolves these proposal-level lane states to predict lane existence,
3D geometry, and visibility:
\begin{equation}
\mathbf{Q}_{i}^{0}
=
\operatorname{MLP}_{q}
\left(
\mathbf{s}_{i}
\oplus
\mathbf{p}_{1:A}
\right),
\qquad
\mathbf{r}_{i}^{0}
=
\sigma
\left(
\operatorname{MLP}_{r}
\left(
\mathbf{Q}_{i}^{0}
\right)
\right).
\end{equation}
Here, $\mathbf{s}_{i}\in\mathbb{R}^{C}$ is the proposal descriptor extracted from
$\mathbf{F}_{c}$, $\mathbf{p}_{1:A}$ are learned anchor embeddings associated with predefined
longitudinal positions, and $\oplus$ denotes anchor-wise broadcasting. The initialized token sequence
$\mathbf{Q}_{i}^{0}$ represents proposal $i$ over ordered lane anchors, while $\mathbf{r}_{i}^{0}$
provides its initial normalized 3D reference state. In this way, each proposal is initialized as a
structured lane state instead of an independent query vector.

At decoder layer $k$, GSD conditions each anchor token on its current 3D reference state. For anchor
$a$ of proposal $i$, the current reference state is converted into a metric 3D point and used to
construct a geometry-conditioned positional embedding:
\begin{equation}
\mathbf{u}_{i,a}^{k}
=
[
x(r_{i,a,x}^{k}),
y_a,
z(r_{i,a,z}^{k})
],
\qquad
\tilde{\mathbf{p}}_{i,a}^{k}
=
\mathbf{p}_{a}
+
\beta
\Phi
\left(
\mathbf{u}_{i,a}^{k}
\right).
\end{equation}
$y_a$ is the fixed longitudinal position of anchor $a$, while $x(\cdot)$ and $z(\cdot)$ map the
normalized reference values to metric lateral and height coordinates. $\Phi(\cdot)$ encodes the
current 3D reference point, and $\beta$ controls the strength of the geometry-conditioned positional
offset. This design makes the query position evolve with the predicted lane geometry instead of
relying only on fixed learned anchor embeddings.

To improve geometry-aware reasoning, each decoder layer contains $M$ independent inquiry branches:
\begin{equation}
(\mathbf{Q}_{m}^{k},\mathbf{h}_{m}^{k})
=
\mathcal{G}_{m}^{k}
\left(
\mathbf{Q}^{k-1}
+
\tilde{\mathbf{P}}^{k},
\mathbf{F}_{c},
\mathbf{r}^{k}
\right),
\qquad
m=1,\ldots,M .
\end{equation}
$\tilde{\mathbf{P}}^{k}$ collects all dynamic positional embeddings
$\tilde{\mathbf{p}}_{i,a}^{k}$. Each branch $\mathcal{G}_{m}^{k}$ contains anchor self-attention,
reference-conditioned cross-attention to $\mathbf{F}_{c}$, and a feed-forward network.
$\mathbf{Q}_{m}^{k}\in\mathbb{R}^{N\times A\times C}$ contains anchor-level states, while
$\mathbf{h}_{m}^{k}\in\mathbb{R}^{N\times C}$ contains lane-level descriptors obtained from the
anchor tokens. The multiple branches provide diverse geometric inquiries over the same lane proposal,
which helps distinguish nearby lanes and maintain coherent lane structures.

The branch outputs are projected and normalized to update the anchor-level state $\mathbf{Q}^{k}$ and
lane-level state $\mathbf{h}^{k}$, which are used for reference refinement and final lane prediction.
After $K$ layers, GSD predicts lane existence, anchor-wise 3D geometry, and visibility while
maintaining proposal identity and ordered lane structure throughout decoding.

\subsection{Loss}

Following LATR~\cite{luo2023latr}, we train RETR with a
weighted combination of a 3D lane loss, a lane proposal
loss, and an auxiliary 2D lane loss:
\begin{equation}
\mathcal{L}_{\mathrm{total}}
=
\mathcal{L}_{3\mathrm{D}}
+
\lambda_{\mathrm{prop}}\mathcal{L}_{\mathrm{prop}}
+
\lambda_{2\mathrm{D}}\mathcal{L}_{2\mathrm{D}},
\end{equation}
where $\lambda_{\mathrm{prop}}=5.0$ and
$\lambda_{2\mathrm{D}}=1.0$.

The 3D lane loss $\mathcal{L}_{3\mathrm{D}}$ supervises
lane existence, anchor-wise lateral and height coordinates,
and visibility. It also includes a projection consistency
term. The proposal loss $\mathcal{L}_{\mathrm{prop}}$
supervises lane proposal generation using classification,
mask, Dice, and objectness terms. The auxiliary 2D loss
$\mathcal{L}_{2\mathrm{D}}$ provides image-plane supervision
through lane segmentation, overlap, and instance embedding.

\section{Experiments}

This work makes the first attempt to study 3D lane detection with event cameras. To validate the effectiveness of our method, we introduce two complementary event-based 3D lane datasets: DSEC-3DLD with real events and Ev-OpenLane with simulated events. Details of the annotation and
dataset generation pipelines are provided in the supplementary material.

\begin{table}[!t]
\centering

% ==========================================================
% Table 1
% ==========================================================
\caption{
Comparison with state-of-the-art 3D lane detection methods on DSEC-3DLD.
We report overall F1-score (F1), precision (P), recall (R), centerline distance error (CD), and
scenario-wise F1-score . Scenario columns denote Straight (Str.), Curved (Cur.),
Intersection (Int.), Merge/Split (M\&S), Normal (Nor.), Overexposed (Ove.),
Underexposed (Und.), and Up/Down road geometry (U/D). Best and second-best results are
shown in \textbf{bold} and \underline{underlined}, respectively.
}
\label{tab:dsec3dld_combined}
\scriptsize
\setlength{\tabcolsep}{2pt}
\renewcommand{\arraystretch}{1.12}
\resizebox{\linewidth}{!}{
\begin{tabular}{lcccc|cccccccc}
\toprule
\multirow{2}{*}{\textbf{Method}}
& \multicolumn{4}{c|}{\textbf{Overall}}
& \multicolumn{8}{c}{\textbf{Scenario-wise F1-Score (\%)}} \\
\cmidrule(lr){2-5} \cmidrule(lr){6-13}
& \textbf{F1$\uparrow$}
& \textbf{P$\uparrow$}
& \textbf{R$\uparrow$}
& \textbf{CD$\downarrow$}
& \textbf{Str.$\uparrow$}
& \textbf{Cur.$\uparrow$}
& \textbf{Int.$\uparrow$}
& \textbf{M\&S$\uparrow$}
& \textbf{Nor.$\uparrow$}
& \textbf{Ove.$\uparrow$}
& \textbf{Und.$\uparrow$}
& \textbf{U/D$\uparrow$} \\
\midrule
PersFormer~\cite{chen2022persformer}
& 50.1 & 48.7 & 51.5 & 0.127
& 51.5 & 48.3 & 48.8 & 58.9 & 53.3 & 31.3 & 41.8 & 34.5 \\

SparsePers~\cite{yao2023sparse}
& 50.2 & 48.1 & 52.3 & 0.134
& 51.6 & 48.8 & 46.1 & 59.2 & 52.2 & 35.2 & 45.4 & 45.6 \\

Anchor3DLane~\cite{huang2023anchor3dlane}
& 54.7 & 57.9 & 51.8 & 0.127
& 56.6 & 53.8 & 51.2 & 50.7 & 57.3 & 28.3 & 49.0 & 45.7 \\

3DLaneDT~\cite{wang20263dlanedt}
& 53.7 & 55.2 & 52.2 & 0.130
& 54.1 & 54.2 & 51.2 & 53.3 & 56.2 & 24.9 & 48.8 & 43.5 \\

Anchor3DLanePP~\cite{huang2024anchor3dlane++}
& 61.8 & 67.1 & 57.4 & 0.121
& 63.8 & 59.9 & 59.3 & 65.6 & 64.4 & 39.7 & 56.2 & 48.6 \\

LATR~\cite{luo2023latr}
& 69.2 & 71.1 & 67.3 & 0.107
& \underline{70.7} & 67.2 & \underline{68.9} & 70.7 & 72.3 & 49.0 & 60.7 & \underline{61.9} \\

DV-3DLane~\cite{luo2024dv}
& \underline{69.4} & \underline{72.5} & 66.6 & \underline{0.105}
& 70.5 & 68.2 & \underline{68.9} & \underline{72.9} & \underline{72.7} & 52.6 & 59.0 & 58.4 \\

MRDALane~\cite{ning2025multi}
& 69.3 & 70.5 & \underline{68.0} & 0.110
& 70.4 & \underline{69.3} & 65.9 & 69.4 & 72.1 & \underline{56.7} & \underline{60.8} & 56.8 \\

\textbf{RETR (ours)}
& \textbf{75.6} & \textbf{76.3} & \textbf{75.0} & \textbf{0.094}
& \textbf{75.8} & \textbf{76.0} & \textbf{73.7} & \textbf{76.8}
& \textbf{77.4} & \textbf{59.6} & \textbf{72.1} & \textbf{71.7} \\
\bottomrule
\end{tabular}
}

% ==========================================================
% Table 2
% ==========================================================
\caption{
Comparison with state-of-the-art 3D lane detection methods on Ev-OpenLane.
We report F1-score (\%), near/far X-error, and near/far Z-error under the standard OpenLane evaluation protocol.
Higher F-score is better, while lower X/Z errors indicate more accurate 3D localization.
Best and second-best results are shown in \textbf{bold} and \underline{underlined}, respectively.
}
\label{tab:openlane_overall}
\scriptsize
\setlength{\tabcolsep}{4.0pt}
\renewcommand{\arraystretch}{1.08}
\resizebox{\linewidth}{!}{
\begin{tabular}{lccccc}
\toprule
\textbf{Method}
& \textbf{F1-score(\%)$\uparrow$}
& \textbf{X-error (near)$\downarrow$}
& \textbf{X-error (far)$\downarrow$}
& \textbf{Z-error (near)$\downarrow$}
& \textbf{Z-error (far)$\downarrow$} \\
\midrule
PersFormer~\cite{chen2022persformer}
& 50.5 & 0.485 & 0.553 & 0.364 & 0.431 \\

Anchor3DLane~\cite{huang2023anchor3dlane}
& 53.1 & 0.300 & 0.311 & 0.103 & 0.139 \\

BEVLaneDet~\cite{wang2023bev}
& 58.4 & 0.309 & 0.659 & 0.244 & 0.631 \\

LaneCPP~\cite{pittner2024lanecpp}
& 60.3 & 0.264 & 0.310 & 0.077 & 0.117 \\

LATR~\cite{luo2023latr}
& 61.9 & \underline{0.219} & 0.259 & 0.075 & \underline{0.104} \\

HeightLane~\cite{park2025heightlane}
& 62.7 & 0.240 & 0.266 & 0.116 & 0.165 \\

RFTR~\cite{li2024repvf}
& 61.8 & 0.341 & 0.450 & \underline{0.073} & 0.107 \\

PVALane~\cite{zheng2024pvalane}
& 62.7 & 0.232 & 0.259 & 0.092 & 0.118 \\

SC-Lane~\cite{park2025sc}
& \underline{64.3} & 0.227 & \underline{0.251} & 0.088 & 0.128 \\

\textbf{RETR (ours)}
& \textbf{66.7} & \textbf{0.203} & \textbf{0.230} & \textbf{0.071} & \textbf{0.101} \\
\bottomrule
\end{tabular}
}
\vspace{-5mm}
\end{table}
% \vspace{-2mm}

\subsection{Datasets Establishment and Evaluation Metrics}

\textbf{Ev-OpenLane} is built upon OpenLane~\cite{chen2022persformer}, a comprehensive large-scale
3D lane detection benchmark based on the Waymo dataset~\cite{sun2020scalability}. It contains
1000 segments and 200K frames captured under diverse weather, terrain, and brightness
conditions at a resolution of $1280 \times 1920$. We augment the original image frames with
synchronized synthetic event data, while preserving 880K lane annotations across 14 categories,
thereby providing a scalable benchmark for 3D lane detection with synchronized images and events.

\textbf{DSEC-3DLD} is built upon  DSEC dataset~\cite{gehrig2021dsec}, a real-world
driving dataset captured with synchronized standard cameras, event cameras, and LiDAR sensors.
Rather than using DSEC only as a source of multimodal recordings, we convert it into a metric
RGB-event 3D lane detection benchmark. The resulting dataset contains synchronized image and event
observations together with 3D lane annotations in autonomous-driving scenes, covering diverse road
layouts, elevation changes, and challenging illumination conditions. This setting allows us to
evaluate whether event sensing can improve 3D lane perception under real sensor noise, motion,
exposure variation, and sparse high-frequency lane evidence.

% DSEC-3DLD contains 30,156 synchronized RGB-event samples
% and 82,746 annotated 3D lane instances from 52 driving
% sequences. We use a sequence-level split comprising
% 33 training, 6 validation, and 13 test sequences.
% All samples from the same sequence remain in one split,
% preventing temporally adjacent observations from being
% shared across training, validation, and test sets.

% To construct 3D lane labels, we first annotate lane structures on the synchronized camera views and
% associate them with metric depth cues derived from the calibrated LiDAR observations. The annotated
% 2D lane traces are lifted into the ego-coordinate system using camera-LiDAR calibration, after which
% we fit ordered 3D lane curves along predefined longitudinal positions. We further refine the lifted
% lanes by checking geometric continuity, road-surface consistency, and image-event alignment, and
% remove ambiguous or severely occluded lane segments that cannot be reliably localized in 3D. This
% annotation pipeline provides lane-level existence labels, anchor-wise 3D coordinates, and visibility
% states for training and evaluation.  Further details are provided in
% Supplementary Material A.
To construct metric 3D lane annotations, we follow a multi-stage annotation and verification
pipeline. We first annotate lane structures on the synchronized camera views, where lane markings
are traced in the image plane with lane-level instance identities. These 2D lane traces are then
associated with metric depth cues derived from calibrated LiDAR observations. Using the
camera-LiDAR calibration, we lift the annotated image-plane lanes into the ego-coordinate system and
fit ordered 3D lane curves along predefined longitudinal anchor positions. To improve annotation
quality, the lifted lanes are further refined through geometric consistency checks, including curve
continuity, road-surface compatibility, camera-LiDAR projection consistency, and image-event
alignment. Lane segments that are highly ambiguous, severely occluded, or cannot be reliably
localized in 3D are removed from the annotations. The resulting annotation pipeline provides
lane-level existence labels, anchor-wise 3D coordinates, and visibility states, which are used for
both training and evaluation. Additional details of the annotation procedure are provided in
Supplementary Material A.

\textbf{Evaluation Metrics.} Ev-OpenLane evaluation protocol computes the F1-score by measuring the Euclidean distance
between uniformly sampled lane points together with a range-IoU criterion. In addition, the
$x$- and $z$-errors are reported in the near range $(0$--$40\,\mathrm{m})$ and far range
$(40$--$100\,\mathrm{m})$ to assess the geometric accuracy of 3D lane predictions \cite{chen2022persformer,guo2020gen}.
DSEC-3DLD follow the distance-based evaluation setting of ONCE-3DLanes \cite{yan2022once}. Predicted
and ground-truth 3D lane curves are matched according to their centerline distance (CD) \cite{yan2022once}, and Precision, Recall and F1-score are computed based on the matched lanes. 

\vspace{-3mm}
\subsection{Implementation Details}

For Ev-OpenLane, all input images are resized to $720 \times 960$, and ResNet-50 \cite{he2016deep} is adopted as
the default backbone. The model is trained for 24 epochs with an initial learning rate of
$2 \times 10^{-4}$. For DSEC-3DLD, we use an input resolution of $480 \times 640$ with 
ResNet-50 \cite{he2016deep} backbone and train for 24 epochs with an initial learning rate of $1 \times 10^{-4}$.
All models are optimized using AdamW \cite{loshchilov2017decoupled} with a weight decay of $0.01$ and gradient clipping of
$20$.  All comparison methods and ablation variants
follow the same training protocol unless otherwise stated.
% 将整个环境移到上一页对应段落之前
\begin{figure}[!t]
    \centering
    \includegraphics[
        width=\linewidth,
        height=0.34\textheight,
        keepaspectratio
    ]{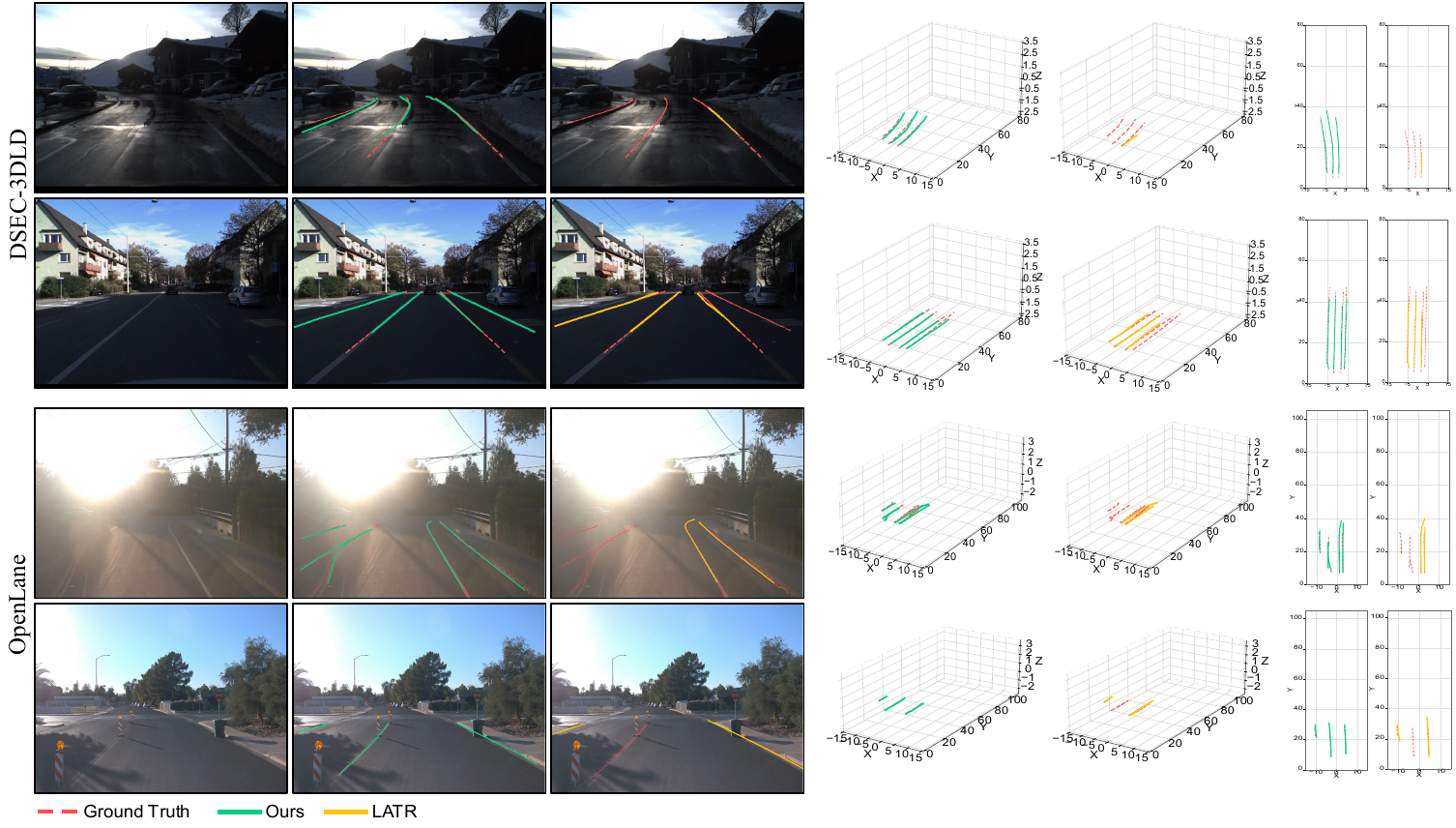}
    \caption{
Qualitative comparison on DSEC-3DLD and Ev-OpenLane.
We compare RETR with the RGB-only LATR baseline under
challenging illumination and road-geometry conditions.
Each example shows image-plane lane predictions on the
left and corresponding 3D geometry and top-view
projections on the right.
Red, green, and yellow denote ground truth, RETR,
and LATR, respectively. Zoom in for details.
    }
    \label{fig:qualitative_results}
    \vspace{-5mm}
\end{figure}
\vspace{-2mm}
\subsection{Main Results}

\textbf{Results on DSEC-3DLD.}
Tab.~\ref{tab:dsec3dld_combined} reports the comparison on DSEC-3DLD. Our approach achieves
the best overall performance, with an F1 score of 75.6\%, precision of 76.3\%, recall of
75.0\%, and the lowest centerline distance of 0.094. Compared with the RGB only LATR
baseline, it improves the overall F1 score by 6.4 points and reduces the centerline distance
from 0.107 to 0.094. The gains are more pronounced under challenging conditions, with
10.6 and 11.4 F1 point improvements in overexposed and underexposed scenes, respectively.
On up and down roads, it further improves the F1 score by 9.8 points, indicating stronger
3D geometric robustness under large height variations. These results show that our reliability
based fusion, structural consolidation, and geometry aware decoding effectively exploit the
event modality for robust and accurate 3D lane detection.

\textbf{Results on Ev-OpenLane.}
Tab.~\ref{tab:openlane_overall} shows the results on Ev-OpenLane under the standard OpenLane
evaluation protocol. Our approach achieves an F1 score of 66.7\%, outperforming the previous
best method by 2.4 points and improving over the RGB only LATR baseline by 4.8 points.
It also achieves the lowest localization errors. Compared with SC-Lane, the near and far
$x$ errors are reduced from 0.227/0.251 to 0.203/0.230, while the near and far $z$ errors
are reduced from 0.088/0.128 to 0.071/0.101. Together with the results on DSEC-3DLD,
these improvements demonstrate the effectiveness and scalability of our approach on
large-scale 3D lane detection benchmarks. Tab.~\ref{tab:supp_evopenlane_scenario} further
reports scenario-wise results.

\textbf{Qualitative analysis.} 
Fig.~\ref{fig:qualitative_results} shows that our approach gives more accurate predictions than the RGB
only LATR baseline under extreme lighting conditions. When strong glare or overexposure makes
lane markings faint in the image, LATR often produces missing or fragmented lanes. Our method
recovers more complete lane boundaries and maintains better consistency in 3D space. The
visual comparisons support the quantitative results, showing that the event modality helps our
model remain reliable when RGB appearance becomes degraded.

\vspace{-3mm}
% \end{table*}

\subsection{Ablation Studies and Analysis}
\begin{wraptable}{r}{0.52\linewidth}
\vspace{-4.5mm}
\centering
\caption{
Comparison of component variants in RETR.
Results are reported as F1-score (\%) on
DSEC-3DLD and Ev-OpenLane.
}
\label{tab:ablation_components}
\scriptsize
\setlength{\tabcolsep}{3.0pt}
\renewcommand{\arraystretch}{0.98}
\resizebox{\linewidth}{!}{
\begin{tabular}{lccccc}
\toprule
\multirow{2}{*}{\textbf{Component}}
& \multirow{2}{*}{\textbf{RCF}}
& \multirow{2}{*}{\textbf{UASC}}
& \multirow{2}{*}{\textbf{GSD}}
& \multicolumn{2}{c}{\textbf{F1-score (\%)$\uparrow$}} \\
\cmidrule(lr){5-6}
& & & & \textbf{DSEC} & \textbf{Ev-Open} \\
\midrule
RGB only
& \xmark & \xmark & \xmark
& 69.2 & 61.9 \\
RCF
& \cmark & \xmark & \xmark
& 72.9 & 64.5 \\
RCF + UASC
& \cmark & \cmark & \xmark
& 74.3 & 65.7 \\
\textbf{Ours}
& \cmark & \cmark & \cmark
& \textbf{75.6} & \textbf{66.7} \\
\bottomrule
\end{tabular}
}
\vspace{-2mm}
\end{wraptable}
\textbf{Component analysis.}
Tabs.~\ref{tab:ablation_components}
and~\ref{tab:ablation_fusion} examine the contributions
of the proposed components and fusion strategy. Compared with the RGB-only LATR baseline, naive fusion
of RGB and event inputs improves the F1-score from 69.2\% to 71.8\% on
 DSEC-3DLD and
from 61.9\% to 63.6\% on Ev-OpenLane, showing that the event modality provides useful
complementary information. Replacing naive fusion with RCF further improves the performance
to 72.9\% and 64.5\%, respectively. Adding UASC brings the F1-score to 74.3\% on DSEC-3DLD
and 65.7\% on Ev-OpenLane, indicating the benefit of preserving structural lane evidence after
multi-scale aggregation. With the full pipeline, our method achieves 75.6\% and 66.7\% F1-score
on the two datasets, outperforming the RGB-only baseline by 6.4 and 4.8 points, respectively.
\sbox{\uasctabbox}{%
\footnotesize
\setlength{\tabcolsep}{1.4pt}
\renewcommand{\arraystretch}{1.02}
\begin{tabular}{lcccc}
\toprule
\multirow{2}{*}{\textbf{Variant}}
& \multirow{2}{*}{\textbf{S\&C}}
& \multirow{2}{*}{\textbf{Gate}}
& \multicolumn{2}{c}{\textbf{F1-score (\%)$\uparrow$}} \\
\cmidrule(lr){4-5}
& & & \textbf{DSEC-3DLD} & \textbf{Ev-OpenLane} \\
\midrule
w/o UASC
& \xmark & \xmark
& 74.1 & 65.2 \\
w/o Gate
& \cmark & \xmark
& 75.1 & 66.3 \\
\textbf{Ours}
& \cmark & \cmark
& \textbf{75.6} & \textbf{66.7} \\
\bottomrule
\end{tabular}
}

\begin{wraptable}[9]{r}{\dimexpr\wd\uasctabbox+0.5em\relax}
\vspace{-4.5mm}
\centering
\caption{Comparison of structural consolidation variants in UASC.}
\label{tab:ablation_uasc}
\usebox{\uasctabbox}
\vspace{-2mm}
\end{wraptable}
\textbf{Effect of structural consolidation.}
Tab.~\ref{tab:ablation_uasc} evaluates the contributions
of structural consolidation and uncertainty-guided gating.
Without UASC, the model achieves F1-scores of 74.1\%
and 65.2\% on DSEC-3DLD and Ev-OpenLane, respectively.
Introducing the structural and contextual branches
without the gate improves these results to 75.1\%
and 66.3\%, yielding gains of 1.0 and 1.1 percentage
points. These improvements support the effectiveness
of combining structural details with multi-scale context.
Adding the uncertainty gate further improves the
F1-scores to 75.6\% and 66.7\%, providing additional
gains of 0.5 and 0.4 percentage points.
The complete UASC therefore improves over the variant
without UASC by 1.5 percentage points on both datasets,
supporting the joint use of structural consolidation
and adaptive refinement.

\begin{table}
\centering
\caption{
Scenario-wise quantitative comparison on the Ev-OpenLane validation set using F1-score.
The best result in each column is highlighted in bold, and the second-best result is underlined.
}
\label{tab:supp_evopenlane_scenario}
\scriptsize
\setlength{\tabcolsep}{4.2pt}
\renewcommand{\arraystretch}{1.08}
\begin{tabular}{lccccccc}
\toprule
\textbf{Method}
& \textbf{All}
& \textbf{Up \& Down}
& \textbf{Curve}
& \textbf{Extreme Weather}
& \textbf{Night}
& \textbf{Intersection}
& \textbf{Merge \& Split} \\
\midrule
BEVLaneDet~\cite{wang2023bev}
& 58.4 & 48.7 & 63.1 & 53.4 & 53.4 & 50.3 & 53.7 \\

LATR~\cite{luo2023latr}
& 61.9 & \underline{55.2} & 68.2 & 57.1 & 55.4 & 52.3 & 61.5 \\

HeightLane~\cite{park2025heightlane}
& 62.7 & 53.6 & 69.3 & 55.4 & 54.6 & 54.1 & 61.1 \\

PVALane~\cite{zheng2024pvalane}
& 62.7 & 54.1 & 67.3 & \textbf{62.0} & \underline{57.2} & 53.4 & 60.0 \\

SC-Lane~\cite{park2025sc}
& \underline{64.3} & 54.6 & \underline{70.7} & 58.1 & 56.5 & \underline{56.1} & \underline{63.0} \\

\textbf{RETR (Ours)}
& \textbf{66.7} & \textbf{60.0} & \textbf{73.1} & \underline{59.8} & \textbf{60.5} & \textbf{58.9} & \textbf{65.9} \\
\bottomrule
\end{tabular}
\end{table}

\begin{table}[t]
\centering

\begin{minipage}[t]{0.29\linewidth}
\vspace{0pt}
\centering
\caption{Effect of structural stages in UASC.}
\label{tab:uasc_stages}
\scriptsize
\setlength{\tabcolsep}{3pt}
\renewcommand{\arraystretch}{1.15}
\begin{tabular*}{\linewidth}{
@{\extracolsep{\fill}}lccc@{}
}
\toprule
$L_h$ & 1 & 2 & 3 \\
\midrule
F1-score (\%) & 74.8 & \textbf{75.6} & 75.0 \\
\bottomrule
\end{tabular*}
\end{minipage}
\hfill
\begin{minipage}[t]{0.37\linewidth}
\vspace{0pt}
\centering
\caption{Effect of refinement cues in UASC.}
\label{tab:uasc_cues}
\scriptsize
\setlength{\tabcolsep}{2pt}
\renewcommand{\arraystretch}{1.15}
\begin{tabular*}{\linewidth}{
@{\extracolsep{\fill}}lcccc@{}
}
\toprule
Cues
& $\mathbf{E}{+}\mathbf{A}$
& $\mathbf{E}{+}\mathbf{D}$
& $\mathbf{A}{+}\mathbf{D}$
& All \\
\midrule
F1-score (\%) & 74.9 & 75.2 & 74.7 & \textbf{75.6} \\
\bottomrule
\end{tabular*}
\end{minipage}
\hfill
\begin{minipage}[t]{0.29\linewidth}
\vspace{0pt}
\centering
\caption{Effect of inquiry branches in GSD.}
\label{tab:gsd_branches}
\scriptsize
\setlength{\tabcolsep}{3pt}
\renewcommand{\arraystretch}{1.15}
\begin{tabular*}{\linewidth}{
@{\extracolsep{\fill}}lccc@{}
}
\toprule
$M$ & 1 & 2 & 3 \\
\midrule
F1-score (\%) & 74.5 & \textbf{75.6} & 75.1 \\
\bottomrule
\end{tabular*}
\end{minipage}

\end{table}
\vspace{-3mm}
\paragraph{Structural stages in UASC.}
Tab.~\ref{tab:uasc_stages} evaluates the number of
high-resolution structural stages $L_h$ on DSEC-3DLD,
with all other settings unchanged. Increasing $L_h$
from one to two improves the F1-score from 74.8\%
to 75.6\%, supporting the use of multi-level structural
features. Increasing $L_h$ to three yields 75.0\%,
with no further improvement. We therefore adopt
$L_h=2$, which achieves the best performance among
the evaluated configurations.
\vspace{-3mm}
\paragraph{Refinement cues in UASC.}
Tab.~\ref{tab:uasc_cues} examines the contributions of
the entropy cue $\mathbf{E}$, high-frequency cue
$\mathbf{A}$, and learned detail cue $\mathbf{D}$.
Removing these cues individually reduces the F1-score
by 0.9, 0.4, and 0.7 percentage points, respectively.
The three cues characterize complementary aspects
of the feature representation:
$\mathbf{E}$ measures ambiguity in the projected
feature distribution, $\mathbf{A}$ captures local
high-frequency variations, and $\mathbf{D}$ provides
learned detail responses.
Together, they allow the refinement gate to account
for both feature ambiguity and local structural evidence.
Removing $\mathbf{E}$ causes the largest performance
drop, from 75.6\% to 74.7\%, highlighting the value
of ambiguity information beyond detail cues alone.
None of the evaluated two-cue combinations matches
the complete configuration, supporting the complementary
roles of all three cues in uncertainty-guided
structural refinement.
\vspace{-3mm}
\paragraph{Inquiry branches in GSD.}
Tab.~\ref{tab:gsd_branches} evaluates the number of
inquiry branches $M$ on DSEC-3DLD. Increasing $M$
from one to two improves the F1-score from 74.5\%
to 75.6\%. The branches independently process the same
lane states through geometry-conditioned queries,
and their outputs are aggregated to update the anchor
representations. The 1.1-percentage-point improvement
supports the effectiveness of combining multiple
geometric inquiries. Increasing $M$ to three yields
75.1\%, without improving on the two-branch configuration.
We therefore use $M=2$ in the default model.

% \begin{wraptable}[6]{r}{0.31\linewidth}
% \vspace{-4mm}
% \centering
% \caption{Ablation of fusion strategies.}
% \label{tab:ablation_fusion}
% \scriptsize
% \setlength{\tabcolsep}{2.2pt}
% \renewcommand{\arraystretch}{0.98}
% \resizebox{\linewidth}{!}{
% \begin{tabular}{lcc}
% \toprule
% \multirow{2}{*}{\textbf{Method}}
% & \multicolumn{2}{c}{\textbf{F1-score (\%)$\uparrow$}} \\
% \cmidrule(lr){2-3}
% & \textbf{DSEC} & \textbf{Ev-Open} \\
% \midrule
% Concat.
% & 71.8 & 63.6 \\
% Cross Attn.
% & 72.3 & 63.4 \\
% CBAM~\cite{woo2018cbam}
% & 72.1 & 63.8 \\
% \textbf{RCF}
% & \textbf{72.9} & \textbf{64.5} \\
% \bottomrule
% \end{tabular}
% }
% \vspace{-4mm}
% \end{wraptable}

\sbox{\fusiontabbox}{%
\footnotesize
\setlength{\tabcolsep}{2.0pt}
\renewcommand{\arraystretch}{1.02}
\begin{tabular}{lcc}
\toprule
\multirow{2}{*}{\textbf{Method}}
& \multicolumn{2}{c}{\textbf{F1-score (\%)$\uparrow$}} \\
\cmidrule(lr){2-3}
& \textbf{DSEC-3DLD} & \textbf{Ev-OpenLane} \\
\midrule
Concatenation
& 71.8 & 63.6 \\
Cross Attention
& 72.3 & 63.4 \\
CBAM~\cite{woo2018cbam}
& 72.1 & 63.8 \\
\textbf{RCF}
& \textbf{72.9} & \textbf{64.5} \\
\bottomrule
\end{tabular}
}
\begin{wraptable}[10]{r}{\dimexpr\wd\fusiontabbox+0.5em\relax}
\vspace{-4.5mm}
\centering
\caption{Comparison of image and event fusion strategies.}
\label{tab:ablation_fusion}
\usebox{\fusiontabbox}
\vspace{-4mm}
\end{wraptable}
\textbf{Effect of fusion strategy.}
Tab.~\ref{tab:ablation_fusion} compares RGB-event fusion strategies under the same backbone and
decoder, isolating the effect of cross-modal interaction. Direct concatenation improves over the
RGB-only baseline, confirming that event observations provide complementary lane cues. However, it
treats RGB and event features as homogeneous channels and does not explicitly address their
appearance-structure discrepancy. Among the evaluated fusion strategies, RCF achieves
the best performance on both benchmarks, improving
over concatenation by 1.1 and 0.9 percentage points
on DSEC-3DLD and Ev-OpenLane, respectively. This validates our
design that effective fusion should first identify modality-consistent lane evidence and then
exchange complementary context bidirectionally.
\vspace{-4mm}
\section{Conclusion}
\vspace{-2mm}
We introduced RETR, a structure preserving transformer for 3D lane detection with RGB and event
inputs. To the best of our knowledge, this work is the first to bring event cameras into 3D lane
detection and to establish dedicated multimodal benchmarks for this setting, including DSEC-3DLD
with real event data and Ev-OpenLane with simulated event observations. Our results show that aligning complementary RGB-event
evidence, preserving structural details, and decoding
ordered lane geometry together improve 3D lane
detection across both benchmarks. RETR follows this principle by aligning
modality-specific evidence through Reciprocal Context Flow Fusion, preserving thin and uncertain
lane structures with Uncertainty Aware Structural Consolidation, and decoding ordered lane
hypotheses with the Geometry State Decoder. Extensive experiments on both benchmarks show that RETR
achieves state-of-the-art performance and strong robustness under challenging illumination and road
geometry, validating the effectiveness of our structure preserving design and demonstrating the
importance of event cameras for robust 3D lane detection.

\begin{ack}
This work is supported by the National Natural Science Foundation of China (62302045) and the Fundamental Research Funds for the Central Universities.
\end{ack}

\bibliographystyle{ieee_fullname}
\bibliography{reference}
%%%%%%%%%%%%%%%%%%%%%%%%%%%%%%%%%%%%%%%%%%%%%%%%%%%%%%%%%%%%
\clearpage
\appendix
\section*{Supplementary Material}
The supplementary material provides dataset construction
and annotation details, event representations, model
implementation details, additional modality and efficiency
analyses, and further qualitative results.
\section{Additional Dataset and Model Details}

\subsection{DSEC-3DLD Statistics and Data Splits}
\label{sec:supp_dsec_statistics}

DSEC-3DLD comprises 52 driving sequences with 30,156
synchronized RGB-event samples and 82,746 annotated
3D lane instances. We partition the dataset at the
sequence level into 33 training, 6 validation, and
13 test sequences, as summarized in
Tab.~\ref{tab:supp_dsec_splits}.
Frames are uniformly sampled at 10 Hz within each
sequence. All samples from a given sequence are assigned
to the same split to prevent cross-split leakage from
temporally adjacent observations within that sequence.

% \begin{table}[htbp]
% \centering
% \caption{Dataset statistics and sequence-level splits
% of DSEC-3DLD.}
% \label{tab:supp_dsec_splits}
% \small
% \begin{tabular}{lrrr}
% \toprule
% Split & Sequences & Samples & Lane instances \\
% \midrule
% Training   & 33 & 19,048 & 52,422 \\
% Validation &  6 &  4,749 & 12,503 \\
% Test       & 13 &  6,359 & 17,821 \\
% \midrule
% Total      & 52 & 30,156 & 82,746 \\
% \bottomrule
% \end{tabular}
% \end{table}

\subsection{DSEC-3DLD Annotation Pipeline}

\begin{figure}[!htbp]
    \centering
    \includegraphics[width=0.98\linewidth]{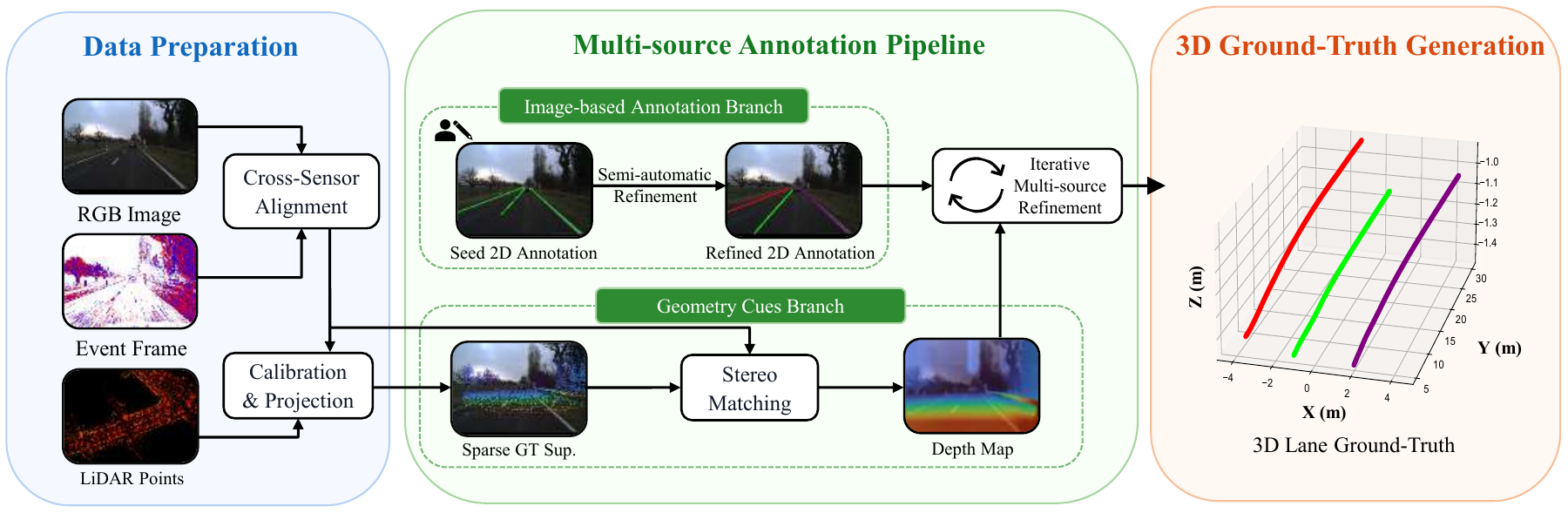}
    \caption{
    \textbf{Annotation pipeline of DSEC-3DLD.}
    We first prepare synchronized RGB images, event frames, and LiDAR points through cross-sensor
    alignment and calibration-based projection. Then, we construct a multi-source annotation pipeline
    with an image-based annotation branch and a geometry-cue branch. The image-based branch provides
    refined 2D lane annotations, while the geometry branch provides sparse geometric supervision and
    depth cues. These complementary sources are jointly refined to generate metric 3D lane
    ground-truth.
    }
    \label{fig:supp_dsec_pipeline}
\end{figure}

DSEC-3DLD is constructed to provide real-world RGB-event data with metric 3D lane annotations.
Since existing event-camera driving datasets do not directly provide 3D lane labels, we build a
multi-source annotation pipeline that combines image evidence, event observations, calibration
information, and geometric cues. The overall pipeline is shown in Fig.~\ref{fig:supp_dsec_pipeline}.
It consists of three main stages: data preparation, multi-source annotation, and 3D ground-truth
generation.

\paragraph{Data preparation.}
For each driving sequence, we first synchronize RGB images and event observations and align them
with the available geometric measurements. The event stream is converted into an event frame that
preserves polarity and motion-induced structural responses. In parallel, LiDAR points are projected
to the camera view using calibrated extrinsic and intrinsic parameters. This cross-sensor alignment
step provides a common image coordinate system for RGB appearance, event responses, and sparse
3D geometry.

More specifically, given a LiDAR point $\mathbf{p}_{l}=[x_l,y_l,z_l,1]^{\top}$ in the LiDAR
coordinate system, we project it into the camera image plane by
\begin{equation}
    \tilde{\mathbf{u}}
    =
    \mathbf{K}
    \mathbf{T}_{c \leftarrow l}
    \mathbf{p}_{l},
    \qquad
    \mathbf{u}
    =
    \left[
    \frac{\tilde{u}_x}{\tilde{u}_z},
    \frac{\tilde{u}_y}{\tilde{u}_z}
    \right]^{\top},
\end{equation}
where $\mathbf{K}$ is the camera intrinsic matrix and $\mathbf{T}_{c \leftarrow l}$ denotes the
rigid transformation from the LiDAR coordinate system to the camera coordinate system. Points
outside the image range or behind the camera are discarded. The projected points provide sparse
metric depth cues for lane regions.

\paragraph{Image-based annotation branch.}
The first branch focuses on image-space lane annotation. Since the RGB camera and event camera
have different imaging planes, viewpoints, and resolutions, we first align the RGB frame and the
event frame into a unified image coordinate system using camera calibration and homography-based
warping~\cite{tomy2022fusing}. We then perform manual initial annotation on the synchronized and spatially aligned RGB
frames to obtain seed 2D lane annotations. These seed annotations define the number of lane
instances, their image-space locations, visible extents, and coarse lane topology.

However, relying only on RGB images can be affected by overexposure, underexposure, shadows,
motion blur, and weak lane textures. Therefore, we further incorporate the aligned event frames for
semi-automatic refinement. Event frames can preserve structural responses along lane boundaries
when RGB appearance is degraded, especially under strong illumination changes or locally weak
textures. By jointly inspecting RGB appearance and event responses, we correct the initial 2D
annotations and obtain more stable refined 2D lane annotations.

% \paragraph{Geometry-cue branch.}
\paragraph{Geometry-cue branch.}
The second branch provides metric geometric constraints for 3D lane annotation. After projecting
LiDAR points onto the image plane using the calibrated LiDAR-to-camera transformation, we obtain
sparse metric depth measurements around the road surface and lane regions. These sparse depth
values are converted into sparse disparity supervision. For a stereo camera with focal length $f$
and baseline $B$, the disparity corresponding to the projected LiDAR depth $d_L^i$ is computed as
\begin{equation}
    s_L^i = \frac{fB}{d_L^i}.
\end{equation}
The sparse disparity map is then constructed by assigning each projected LiDAR point to its image
location:
\begin{equation}
    \mathbf{S}_L(\mathbf{u}^i) = s_L^i,
\end{equation}
where $\mathbf{u}^i$ denotes the projected image coordinate of the $i$-th LiDAR point.

Using this sparse disparity map as geometric supervision, we employ a state-of-the-art open-source
event-based stereo matching algorithm~\cite{bartolomei2024lidar} to estimate a dense disparity map:
\begin{equation}
    \hat{\mathbf{S}}
    =
    \mathcal{M}_{\mathrm{stereo}}
    \left(
    \mathbf{I}_{L},
    \mathbf{I}_{R},
    \mathbf{E}_{L},
    \mathbf{E}_{R},
    \mathbf{S}_{L}
    \right),
\end{equation}
where $\mathbf{I}_{L}$ and $\mathbf{I}_{R}$ are the stereo RGB frames, $\mathbf{E}_{L}$ and
$\mathbf{E}_{R}$ are the corresponding event representations, $\mathbf{S}_{L}$ is the sparse
LiDAR-derived disparity supervision, and $\hat{\mathbf{S}}$ is the estimated dense disparity map.
The dense depth map is recovered by
\begin{equation}
    \hat{\mathbf{D}}(\mathbf{u})
    =
    \frac{fB}{\hat{\mathbf{S}}(\mathbf{u})+\epsilon},
\end{equation}
where $\epsilon$ is a small constant for numerical stability.

The estimated dense depth map is used as a geometric prior rather than directly as the final
annotation. We sample depth values along each manually refined 2D lane and lift the lane points into
metric 3D space using the calibrated camera model. The lifted lanes are further checked with the
LiDAR projection and manually inspected in both the image view and the ground-coordinate view.
This process converts sparse LiDAR supervision and dense event-stereo geometry into reliable 3D
lane ground-truth.

\paragraph{Iterative multi-source refinement.}
After obtaining the refined 2D lane annotations and dense depth maps, we perform iterative
multi-source refinement by jointly checking RGB appearance, event responses, LiDAR projections,
and stereo depth estimates. This step verifies both image-space lane localization and metric 3D
geometry consistency.

During the generation of 3D labels, we observe that directly fitting 2D lane curves or depth
sequences in the image coordinate system with a global high-order function can introduce severe
oscillatory artifacts, known as the Runge phenomenon. This issue is particularly evident in the
far range, where lane markings are highly compressed in the image plane and small pixel-level
errors can be amplified into large 3D localization errors. In addition, depth candidates around
lane regions are not perfectly continuous, and global high-order interpolation tends to produce
unstable geometry in sparse or noisy regions.

To address this problem, we do not generate final 3D labels by fitting lanes in the image plane.
Instead, we first lift each refined 2D lane point into the camera coordinate system using its
corresponding depth. For an image-space lane point $\mathbf{q}_j=[u_j,v_j]^\top$ with depth
$\hat{d}_j$, the corresponding 3D point in the camera coordinate system is computed as
\begin{equation}
    \mathbf{x}_{c}^{j}
    =
    \hat{d}_{j}
    \mathbf{K}^{-1}
    \left[
    u_j, v_j, 1
    \right]^{\top},
\end{equation}
where $\mathbf{K}$ is the camera intrinsic matrix. The point is then transformed into the
ground-coordinate system by
\begin{equation}
    \mathbf{x}_{g}^{j}
    =
    \mathbf{T}_{g \leftarrow c}
    \begin{bmatrix}
    \mathbf{x}_{c}^{j} \\
    1
    \end{bmatrix}.
\end{equation}
Thus, each lifted lane is represented as a metric 3D point sequence
\begin{equation}
    \hat{\mathcal{L}}
    =
    \left\{
    (\hat{x}_j,\hat{y}_j,\hat{z}_j)
    \right\}_{j=1}^{N}.
\end{equation}

We then perform lane-wise refinement in the metric ground-coordinate system rather than global
high-order fitting in the image plane. Specifically, we use the longitudinal coordinate
$\hat{y}$ as the independent variable and locally smooth or piecewise interpolate the lateral
position $\hat{x}$ and height $\hat{z}$. The refined lane is finally resampled at predefined
longitudinal positions:
\begin{equation}
    \mathcal{L}^{3D}
    =
    \left\{
    \left(
    x(y_k), y_k, z(y_k), m_k
    \right)
    \right\}_{k=1}^{K},
    \qquad
    y_k \in \mathcal{Y},
\end{equation}
where $\mathcal{Y}$ denotes the predefined longitudinal sampling positions and $m_k$ is the
validity indicator. By lifting lane points from the image coordinate system to the camera
coordinate system, transforming them into the ground-coordinate system, and refining them in metric
space, we avoid the Runge phenomenon caused by image-space global fitting and obtain more stable
3D lane annotations with plausible lateral position, height variation, and longitudinal continuity.

\paragraph{Quality control.}
We perform quality control to ensure reliable 3D lane annotations. For scenario categorization,
we first use GPT-4o to assist the preliminary classification of driving scenes, including normal,
overexposed, underexposed, curve, and merge/split cases. The generated scenario labels are then
manually reviewed, and samples with inaccurate or ambiguous labels are reassigned to the correct
categories.

\paragraph{Independent geometric validation.}
We evaluate 400 frames containing 1,126 annotated
lane instances using synchronized geometric measurements
withheld from annotation generation and refinement.
Depth consistency error (DCE) is the absolute difference
between annotated anchor depth and the corresponding
independent measurement, excluding out-of-image anchors
and invalid measurements.
The mean, median, and 95th-percentile DCE are
0.064 m, 0.041 m, and 0.157 m, respectively,
supporting the depth consistency of the annotations.

To further assess the reliability of DSEC-3DLD annotations, we introduce an expert scoring
protocol. We invite ten expert annotators to evaluate the annotated samples from both image-space
and metric 3D perspectives. As shown in Tab.~\ref{tab:supp_expert_score}, the assessment focuses
on three key aspects: 2D annotation accuracy, 3D lane accuracy, and scenario label accuracy. Each
criterion is scored on a 1--5 scale, where a higher score indicates better annotation quality.
Samples with low scores are corrected or removed from the final dataset. The final annotations are
retained only when the image-space lane structure, metric 3D geometry, and scenario labels are all
verified.

% \begin{table}[t]
% \centering
% \caption{
% Expert assessment criteria for DSEC-3DLD annotation quality.
% Each item is scored from 1 to 5, where higher scores indicate better quality.
% }
% \label{tab:supp_expert_score}
% \scriptsize
% \setlength{\tabcolsep}{5pt}
% \renewcommand{\arraystretch}{1.08}
% \begin{tabular}{lp{0.62\linewidth}}
% \toprule
% \textbf{Criterion} & \textbf{Assessment content} \\
% \midrule
% 2D annotation accuracy
% & Whether the annotated 2D lane positions, visible ranges, and lane instances are correct. \\

% 3D lane accuracy
% & Whether the lifted 3D lanes have reasonable lateral position, height, and longitudinal continuity. \\

% Scenario label accuracy
% & Whether the scene categories, such as normal, overexposed, underexposed, curve, and merge/split, are correctly assigned. \\
% \bottomrule
% \end{tabular}
% \end{table}

\subsection{Ev-OpenLane Generation}

OpenLane is a large-scale 3D lane detection benchmark, but it does not provide event-camera
observations. Therefore, it cannot directly support the study of RGB-event multimodal 3D lane
detection. To construct a scalable RGB-event 3D lane benchmark, we generate event observations from
OpenLane using ESIM~\cite{rebecq2018esim}, a standard open-source event camera simulator, while
preserving the original 3D lane annotations.

Directly simulating events from the original OpenLane frames is insufficient because the original
sequences have a relatively low frame rate. Large inter-frame displacement may cause abrupt
brightness changes and unstable event generation. To address this issue, we first resize the images
to $640\times480$ and temporally upsample the RGB sequences to 30 fps by frame interpolation. The
interpolated frames have a temporal interval of 33.33 ms and are then used as the input to ESIM.

Given the interpolated image sequence, ESIM generates events according to the log-intensity change
at each pixel. An event is triggered when the brightness change reaches the positive or negative
contrast threshold:
\begin{equation}
    \Delta L(\mathbf{x}, t)
    =
    L(\mathbf{x}, t)
    -
    L(\mathbf{x}, t_{\mathrm{last}}),
    \qquad
    |\Delta L(\mathbf{x}, t)| \geq C,
\end{equation}
where $L(\mathbf{x}, t)$ denotes the log intensity at pixel $\mathbf{x}$ and time $t$,
$t_{\mathrm{last}}$ is the timestamp of the last event at the same pixel, and $C$ is the contrast
threshold. The event polarity is determined by the sign of $\Delta L(\mathbf{x}, t)$.

We generate the timestamp file required by ESIM according to the 30 fps interpolated frame rate.
The main simulation settings are summarized in Tab.~\ref{tab:supp_esim_params}. The generated event
streams are synchronized with the corresponding OpenLane key frames and paired with the original
3D lane annotations, resulting in Ev-OpenLane for large-scale RGB-event 3D lane detection.

By providing complementary real-world and large-scale simulated RGB-event 3D lane benchmarks, we hope DSEC-3DLD and Ev-OpenLane can serve as useful resources for advancing event-camera-based perception in autonomous driving.
% \vspace{-2mm}
\paragraph{Transfer to real events.}
Training on Ev-OpenLane and evaluating directly on
DSEC-3DLD yields an F1-score of 56.4\%.
Adapting the pretrained model with 10\% of the
DSEC-3DLD training data improves this to 69.4\%,
while full DSEC-3DLD training achieves 75.6\%.
Limited real-event supervision thus substantially
improves transfer performance.
The remaining gap reflects cross-dataset differences,
including sensing conditions and scene distributions,
rather than event simulation alone.

\begin{table}[t]
\centering
\caption{
ESIM simulation settings for generating Ev-OpenLane.
OpenLane images are resized to $640\times480$ and temporally upsampled to 30 fps before event
simulation.
}
\label{tab:supp_esim_params}
\scriptsize
\setlength{\tabcolsep}{4pt}
\renewcommand{\arraystretch}{1.08}
\begin{tabular}{lll}
\toprule
\textbf{Parameter} & \textbf{Setting} & \textbf{Description} \\
\midrule
Original frame rate
& 10 fps
& Frame rate of the original OpenLane key frames. \\

Interpolated frame rate
& 30 fps
& Frame rate after temporal interpolation. \\

Interpolated frame interval
& 33.33 ms
& Temporal interval between adjacent interpolated frames. \\

Image resolution
& $640 \times 480$
& Resolution used for ESIM simulation and event representation. \\

Input timestamp interval
& 33.33 ms
& Timestamp interval for ESIM input frames. \\

Data source
& Image folder
& ESIM reads interpolated frames and timestamps from an image sequence. \\

ESIM data source flag
& \texttt{data\_source=2}
& Image-folder mode in ESIM. \\

Log image
& \texttt{use\_log\_image=1}
& Converts input images into the log-intensity domain. \\

Log epsilon
& \texttt{log\_eps=0.1}
& Numerical offset used in log-intensity conversion. \\

Positive contrast threshold $C^{+}$
& 0.15
& Threshold for generating positive events. \\

Negative contrast threshold $C^{-}$
& 0.15
& Threshold for generating negative events. \\

ESIM publisher frame rate
& 30 fps
& Frame rate used to publish the interpolated input stream. \\

Exposure time
& 10.0 ms
& Exposure time of the simulated APS frames. \\

Event accumulation window
& 100 ms
& Events are accumulated between adjacent OpenLane key frames. \\

% Output representation
% & 2-channel polarity frame
% & Positive and negative events are accumulated separately. \\
\bottomrule
\end{tabular}
\end{table}

\subsection{Event Representation and Accumulation Window}
\label{sec:supp_event_representation}

For DSEC-3DLD, an RGB frame timestamped at $t_k$ is
paired with all valid events in the causal interval
$[t_k-50\,\mathrm{ms},t_k]$. We spatially align the events
with the RGB image and encode them into a
polarity-preserving voxel representation. We do not
manually select lane-related events, and events after
$t_k$ are excluded. The default accumulation window
is 50 ms for DSEC-3DLD and 100 ms for Ev-OpenLane.

\paragraph{Accumulation-window ablation.}
Tab.~\ref{tab:supp_event_window} evaluates the event
accumulation duration on DSEC-3DLD while keeping the
remaining settings unchanged. The 50-ms window achieves
the highest F1-score among the tested durations.
Reducing the window to 25 ms or increasing it to 100 ms
decreases the F1-score by 0.7 and 0.6 percentage points,
respectively.

\subsection{GSD Implementation Details}
\label{sec:supp_gsd_details}

\paragraph{Proposal descriptor extraction.}
The instance activation map (IAM) head predicts $K_s$
activation maps for each lane proposal. After sigmoid
activation and spatial normalization, each map performs
weighted pooling over the consolidated feature
$\mathbf{F}_c$. The pooled descriptors are concatenated
and projected to obtain the proposal descriptor:
\begin{equation}
\mathbf{s}_i =
G\left(
W_s
\left[
\operatorname{Pool}_{i,1}(\mathbf{F}_c);
\ldots;
\operatorname{Pool}_{i,K_s}(\mathbf{F}_c)
\right]
+\mathbf{b}_s
\right),
\end{equation}
where $\operatorname{Pool}_{i,j}$ denotes weighted
spatial pooling using the $j$-th normalized activation
map of proposal $i$. The semicolon denotes feature
concatenation, $W_s$ and $\mathbf{b}_s$ are projection
parameters, and $G$ is the proposal-refinement mapping.
The resulting descriptor
$\mathbf{s}_i\in\mathbb{R}^{C}$ is used to initialize
the ordered anchor tokens of proposal $i$.

\paragraph{Ordered anchor initialization.}
The learned anchor embeddings are implemented as a matrix
$\mathbf{P}=[\mathbf{p}_1;\ldots;\mathbf{p}_A]
\in\mathbb{R}^{A\times C}$ using \texttt{nn.Embedding}
with its default initialization. They are shared across
proposals and jointly optimized with the network.
Each embedding is associated with a predefined
longitudinal anchor position. The initial token for
anchor $a$ of proposal $i$ is constructed as
\begin{equation}
\mathbf{q}_{i,a}^{0}
=
\operatorname{MLP}_{q}
\left(\mathbf{s}_i+\mathbf{p}_a\right).
\end{equation}
Here, $\mathbf{s}_i$ is broadcast across the $A$ anchors
and added to their respective embeddings.
The resulting tokens form
$\mathbf{Q}_i^0\in\mathbb{R}^{A\times C}$.

\paragraph{Inquiry-branch aggregation.}
At decoder layer $k$, let
$\mathbf{q}_{i,a}^{k,m}\in\mathbb{R}^{C}$ denote
the state of anchor $a$ in proposal $i$ from the
$m$-th branch output $\mathbf{Q}_m^k$.
The states from the $M$ branches are concatenated,
projected back to dimension $C$, and normalized:
\begin{equation}
\mathbf{q}_{i,a}^{k}
=
\operatorname{LN}
\left(
W_{\mathrm{agg}}
\left[
\mathbf{q}_{i,a}^{k,1};
\ldots;
\mathbf{q}_{i,a}^{k,M}
\right]
+\mathbf{b}_{\mathrm{agg}}
\right),
\end{equation}
where $W_{\mathrm{agg}}\in\mathbb{R}^{C\times MC}$,
$\mathbf{b}_{\mathrm{agg}}\in\mathbb{R}^{C}$, and
$\operatorname{LN}$ denotes layer normalization.
The aggregated anchor states form
$\mathbf{Q}^{k}\in\mathbb{R}^{N\times A\times C}$.

\paragraph{Default configuration.}
Tab.~\ref{tab:supp_architecture_settings} summarizes
the default architectural settings of RETR.
In addition to the GSD parameters, we report the number
of high-resolution structural stages $L_h$ used by UASC.

\begin{table}[t]
\centering
\captionsetup{
  font=small,
  format=plain,
  labelsep=colon,
  justification=justified,
  singlelinecheck=false,
  aboveskip=0pt,
  belowskip=0pt
}
\setlength{\tabcolsep}{0pt}
\renewcommand{\arraystretch}{0}

\begin{tabular}{@{}c@{\hspace{0.02\linewidth}}c@{}}

% ==================== Captions ====================
\begin{minipage}[t]{0.47\linewidth}
\vspace{0pt}
\caption{
Dataset statistics and sequence-level splits
of DSEC-3DLD.
}
\label{tab:supp_dsec_splits}
\end{minipage}
&
\begin{minipage}[t]{0.51\linewidth}
\vspace{0pt}
\caption{
Default architectural settings of RETR,
including UASC and GSD.
}
\label{tab:supp_architecture_settings}
\end{minipage}
\\[3pt]

% ==================== Left table ====================
\begin{minipage}[t]{0.47\linewidth}
\vspace{0pt}
\centering
\footnotesize
\setlength{\tabcolsep}{3pt}
\renewcommand{\arraystretch}{1.12}
\begin{tabular*}{\linewidth}{
@{\extracolsep{\fill}}lrrr@{}
}
\toprule
\textbf{Split}
& \textbf{Sequences}
& \textbf{Samples}
& \shortstack{\textbf{Lane}\\\textbf{instances}} \\
\midrule
Training   & 33 & 19,048 & 52,422 \\
Validation &  6 &  4,749 & 12,503 \\
Test       & 13 &  6,359 & 17,821 \\
\midrule
Total      & 52 & 30,156 & 82,746 \\
\bottomrule
\end{tabular*}
\end{minipage}
&
% ==================== Right table ====================
\begin{minipage}[t]{0.51\linewidth}
\vspace{0pt}
\centering
\footnotesize
\setlength{\tabcolsep}{4pt}
\renewcommand{\arraystretch}{1.12}
\begin{tabularx}{\linewidth}{
@{}l>{\raggedright\arraybackslash}Xr@{}
}
\toprule
\textbf{Symbol}
& \textbf{Description}
& \textbf{Value} \\
\midrule
$L_h$ & High-resolution structural stages in UASC & 2 \\
$N$   & Initial lane proposals & 12 \\
$K_s$ & Activation maps per proposal & 4 \\
$A$   & Ordered anchors per proposal & 20 \\
$C$   & Hidden dimension & 256 \\
$K$   & Decoder layers & 6 \\
$M$   & Inquiry branches per decoder layer & 2 \\
\bottomrule
\end{tabularx}
\end{minipage}

\end{tabular}
\end{table}

% \begin{table}[t]
% \centering
% \caption{Default architectural settings of RETR.}
% \label{tab:supp_architecture_settings}
% \scriptsize
% \setlength{\tabcolsep}{5pt}
% \renewcommand{\arraystretch}{1.08}
% \begin{tabular}{llr}
% \toprule
% \textbf{Symbol} & \textbf{Description} & \textbf{Value} \\
% \midrule
% $L_h$ & High-resolution structural stages in UASC & 2 \\
% $N$   & Initial lane proposals & 12 \\
% $K_s$ & Activation maps per proposal & 4 \\
% $A$   & Ordered anchors per proposal & 20 \\
% $C$   & Hidden dimension & 256 \\
% $K$   & Decoder layers & 6 \\
% $M$   & Inquiry branches per decoder layer & 2 \\
% \bottomrule
% \end{tabular}
% \end{table}

\subsection{Architecture and Event-Modality Contributions}
\label{sec:supp_modality_controls}

We compare RETR with RGB-only, event-only, and
event-perturbed variants on DSEC-3DLD.
Tab.~\ref{tab:supp_modality_controls} summarizes the
results, with LATR serving as an RGB-only baseline.

\begin{table}[t]
\centering
\captionsetup{
  font=small,
  format=plain,
  labelsep=colon,
  justification=justified,
  singlelinecheck=false,
  aboveskip=0pt,
  belowskip=0pt
}
\setlength{\tabcolsep}{0pt}
\renewcommand{\arraystretch}{0}

\begin{tabular}{@{}c@{\hspace{0.02\linewidth}}
                  c@{\hspace{0.02\linewidth}}c@{}}

% ==================== Top-aligned captions ====================
\begin{minipage}[t]{0.31\linewidth}
\vspace{0pt}
\caption{
Expert assessment of DSEC-3DLD annotations,
reporting mean scores (1--5) and quality-control
pass rates.
}
\label{tab:supp_expert_score}
\end{minipage}
&
\begin{minipage}[t]{0.43\linewidth}
\vspace{0pt}
\caption{
Comparison of RETR input variants on DSEC-3DLD,
including RGB-only, event-only, RGB-event,
and noisy-event configurations.
}
\label{tab:supp_modality_controls}
\end{minipage}
&
\begin{minipage}[t]{0.22\linewidth}
\vspace{0pt}
\caption{
Event window ablation on DSEC-3DLD
with other settings fixed.
}
\label{tab:supp_event_window}
\end{minipage}
\\[2pt]

% ==================== Left table ====================
\begin{minipage}[t]{0.31\linewidth}
\vspace{0pt}
\centering
\footnotesize
\setlength{\tabcolsep}{2pt}
\renewcommand{\arraystretch}{1.10}
\begin{tabular*}{\linewidth}{
@{\extracolsep{\fill}}lrr@{}
}
\toprule
\textbf{Criterion}
& \textbf{Score}
& \textbf{Pass (\%)} \\
\midrule
2D annotation  & 4.62 & 96.4 \\
3D lane        & 4.41 & 93.7 \\
Scenario label & 4.70 & 97.8 \\
\midrule
Overall        & 4.57 & 95.9 \\
\bottomrule
\end{tabular*}
\end{minipage}
&
% ==================== Middle table ====================
\begin{minipage}[t]{0.43\linewidth}
\vspace{0pt}
\centering
\footnotesize
\setlength{\tabcolsep}{2pt}
\renewcommand{\arraystretch}{1.10}
\begin{tabular*}{\linewidth}{
@{\extracolsep{\fill}}llr@{}
}
\toprule
\textbf{Method}
& \textbf{Input}
& \textbf{F1-score (\%)} \\
\midrule
LATR~\cite{luo2023latr}
& RGB & 69.2 \\
RETR-RGB
& RGB & 72.5 \\
RETR-Event
& Event & 64.8 \\
RETR-Noisy
& RGB + noisy event & 74.6 \\
RETR
& RGB + event & \textbf{75.6} \\
\bottomrule
\end{tabular*}
\end{minipage}
&
% ==================== Right table ====================
\begin{minipage}[t]{0.22\linewidth}
\vspace{0pt}
\centering
\footnotesize
\setlength{\tabcolsep}{2pt}
\renewcommand{\arraystretch}{1.10}
\begin{tabular*}{\linewidth}{
@{\extracolsep{\fill}}cr@{}
}
\toprule
\textbf{Accum. window}
& \textbf{F1 (\%)} \\
\midrule
25 ms  & 74.9 \\
50 ms  & \textbf{75.6} \\
100 ms & 75.0 \\
\bottomrule
\end{tabular*}
\end{minipage}

\end{tabular}
\end{table}

\paragraph{Architectural contribution.}
RETR-RGB improves the F1-score from 69.2\% to 72.5\%
compared with LATR, yielding a gain of 3.3 percentage
points without event input. This result demonstrates
an improvement under the RGB-only setting.

\paragraph{Complementary modalities.}
RETR-Event achieves an F1-score of 64.8\%, compared
with 72.5\% for RETR-RGB. Combining RGB and event
observations achieves 75.6\%, outperforming both
single-modality variants. These results support the
use of complementary RGB and event observations
for 3D lane detection.

\paragraph{Sensitivity to event perturbations.}
For RETR-Noisy, we retrain the model with zero-mean
Gaussian noise applied only to active event-voxel
entries, leaving zero-event regions unchanged.
The F1-score decreases from 75.6\% to 74.6\%.
This result indicates sensitivity to event perturbations
and is consistent with a benefit from reliable event
information. Since the perturbation retains part of
the event structure, it does not constitute a complete
removal of event information.

\subsection{Accuracy--Efficiency Tradeoff}
\label{sec:supp_efficiency}

All variants are profiled using NVIDIA RTX 4090 hardware.
Network latency is averaged over 1,000 runs after
100 warm-up iterations, with CUDA synchronization.
Tab.~\ref{tab:supp_efficiency} reports parameter counts,
multiply--accumulate operations (MACs), preprocessing
time, network latency, and F1-score.

\begin{table}[t]
\centering
\caption{
Accuracy and computational cost of RETR and its
single-modality variants on DSEC-3DLD.
Preprocessing time and network latency are reported
separately.
}
\label{tab:supp_efficiency}
\scriptsize
\setlength{\tabcolsep}{4pt}
\renewcommand{\arraystretch}{1.08}
\begin{tabular}{lrrrrr}
\toprule
\textbf{Method}
& \textbf{Params (M)}
& \textbf{MACs (G)}
& \textbf{Preproc. (ms)}
& \textbf{Network (ms)}
& \textbf{F1-score (\%)} \\
\midrule
RETR-RGB
& 83.56 & 117.6 & 69.9 & 65.03 & 72.5 \\
RETR-Event
& 84.93 & 119.2 & 78.5 & 64.92 & 64.8 \\
RETR
& 95.47 & 148.6 & 80.9 & 81.26 & \textbf{75.6} \\
\bottomrule
\end{tabular}
\end{table}

Compared with RETR-RGB, the full RGB-event model improves
the F1-score by 3.1 percentage points, with 11.91M
additional parameters and 31.0G additional MACs.
Preprocessing time increases from 69.9 ms to 80.9 ms,
while network latency increases from 65.03 ms to
81.26 ms. RETR-Event has a similar parameter count
and network latency to RETR-RGB but achieves a lower
F1-score. With 11.91M additional parameters compared with RETR-RGB,
the full model achieves the best F1-score of 75.6\%
among the evaluated variants.

\section{More Qualitative Results and Analysis}

In this section, we present additional qualitative results to provide a more comprehensive evaluation of our method. The visualizations show that our model maintains strong performance across diverse illumination conditions and road structures, including overexposed and underexposed scenes, straight and curved roads, as well as complex areas such as intersections and merge-split regions. These results further verify the advantage of our method in challenging scenarios, and also suggest that event cameras are a promising sensing modality for robust 3D lane detection in autonomous driving.

\paragraph{Broader Impacts}
This work supports robust autonomous-driving perception by advancing RGB-event 3D lane detection and providing benchmarks for event-based multimodal road-structure understanding. These resources may facilitate research under challenging illumination, high dynamic range, and fast-motion conditions. 

\paragraph{Limitation and future work.}
As RGB-Event 3D lane detection is still at an early stage, the availability of real event camera data with metric 3D lane annotations is naturally limited. Our work establishes a first foundation for this setting through dedicated benchmarks and a structure-preserving model. Future research can build upon this foundation by scaling real-world annotations, covering more sensor configurations, and improving deployment efficiency.

\begin{figure*}[t]
    \centering
    \includegraphics[width=\linewidth]{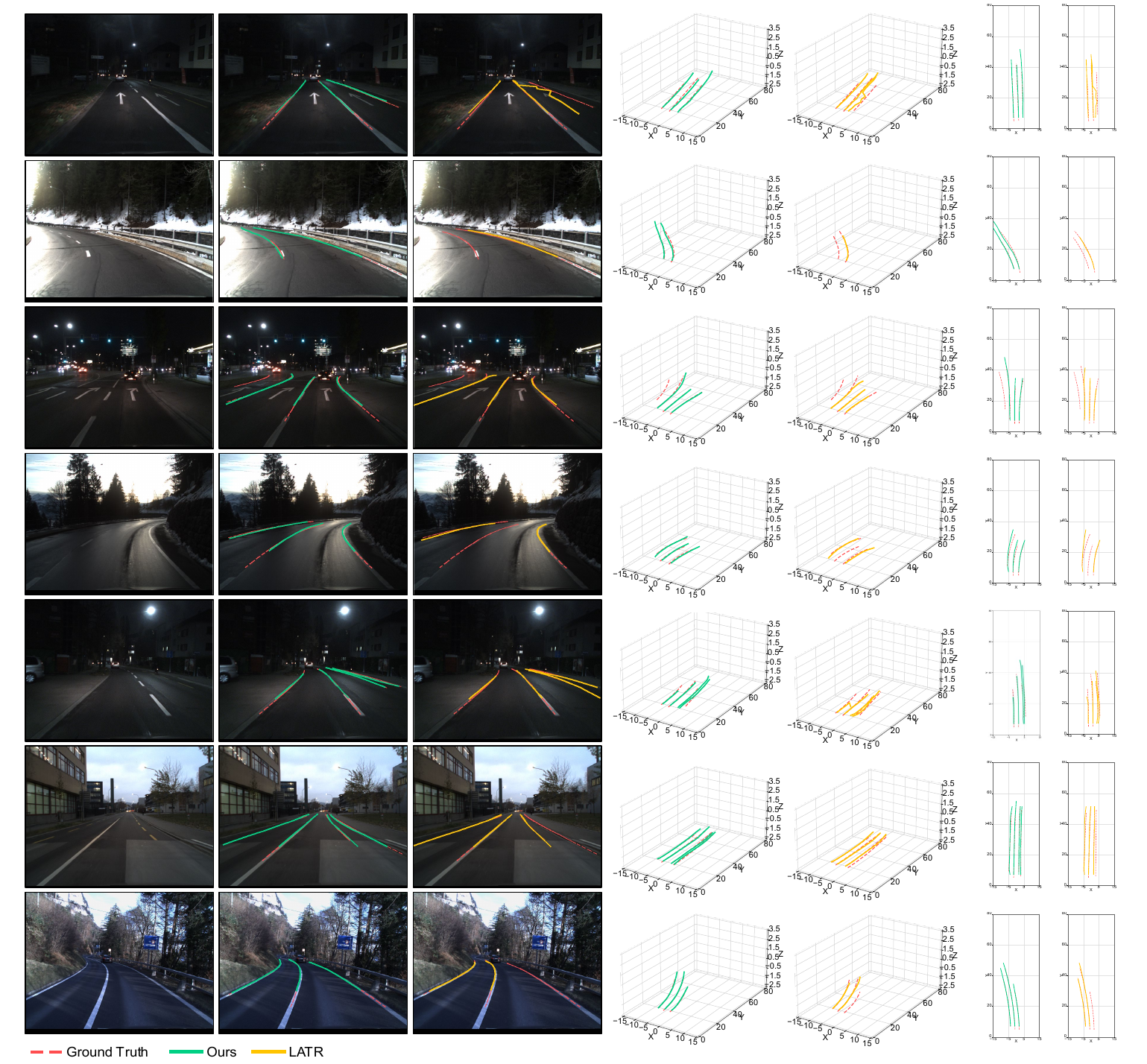}
    \caption{
    Additional qualitative results on DSEC-3DLD. 
    Red dashed lines denote ground truth, green lines denote our predictions, and yellow lines denote LATR predictions. 
    Our method shows better agreement with the ground truth in both image-space and 3D visualizations, demonstrating robust performance under real-world illumination and road-geometry variations.
    }
    \label{fig:supp_dsec_qualitative}
\end{figure*}

\begin{figure*}[t]
    \centering
    \includegraphics[width=\linewidth]{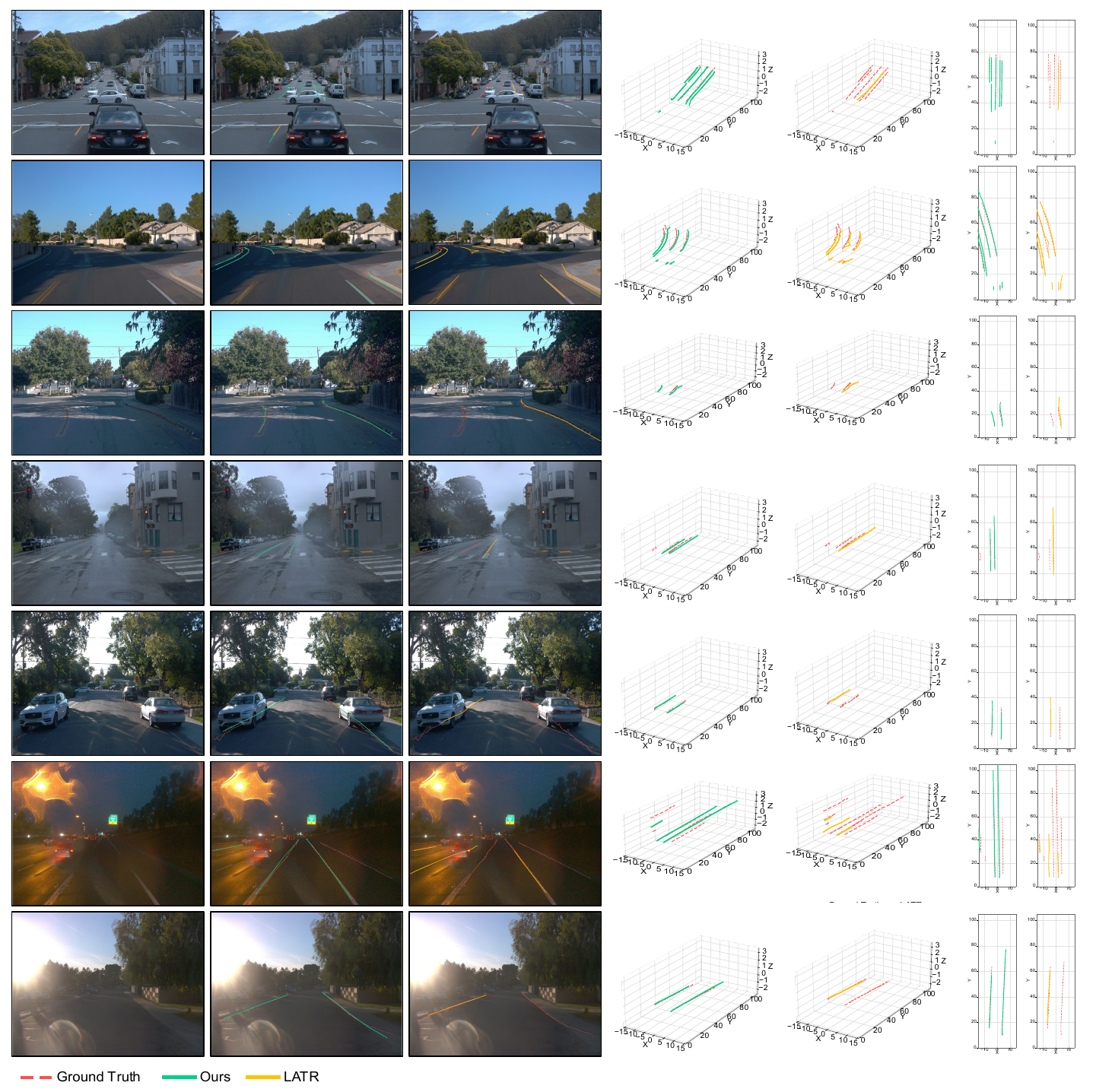}
\caption{
Additional qualitative results on Ev-OpenLane. 
Red dashed lines denote ground truth, green lines denote our predictions, and yellow lines denote LATR predictions. 
Our method produces more coherent image-space and 3D lane predictions than LATR across diverse large-scale scenes, showing strong generalization on simulated RGB-event data.
}
    \label{fig:supp_evopenlane_qualitative}
\end{figure*}

\clearpage
\section*{NeurIPS Paper Checklist}

\begin{enumerate}

\item {\bf Claims}
    \item[] Question: Do the main claims made in the abstract and introduction accurately reflect the paper's contributions and scope?
    \item[] Answer: \answerYes{} % Replace by \answerYes{}, \answerNo{}, or \answerNA{}.
    \item[] Justification: The claims in the abstract and introduction are consistent with the contributions and experimental results.

\item {\bf Limitations}
    \item[] Question: Does the paper discuss the limitations of the work performed by the authors?
    \item[] Answer: \answerYes{} % Replace by \answerYes{}, \answerNo{}, or \answerNA{}.
    \item[] Justification: The paper discusses limitations in the future work section of the supplementary material.

\item {\bf Theory assumptions and proofs}
    \item[] Question: For each theoretical result, does the paper provide the full set of assumptions and a complete (and correct) proof?
    \item[] Answer: \answerNA{} % Replace by \answerYes{}, \answerNo{}, or \answerNA{}.
    \item[] Justification: This work does not include theoretical results.

    \item {\bf Experimental result reproducibility}
    \item[] Question: Does the paper fully disclose all the information needed to reproduce the main experimental results of the paper to the extent that it affects the main claims and/or conclusions of the paper (regardless of whether the code and data are provided or not)?
    \item[] Answer: \answerYes{} % Replace by \answerYes{}, \answerNo{}, or \answerNA{}.
    \item[] Justification: The paper provides experimental details in the main paper and supplementary material.

\item {\bf Open access to data and code}
    \item[] Question: Does the paper provide open access to the data and code, with sufficient instructions to faithfully reproduce the main experimental results, as described in supplemental material?
    \item[] Answer: \answerYes{} % Replace by \answerYes{}, \answerNo{}, or \answerNA{}.
    \item[] Justification:  If the paper is accepted, the code and data will be released. 

\item {\bf Experimental setting/details}
    \item[] Question: Does the paper specify all the training and test details (e.g., data splits, hyperparameters, how they were chosen, type of optimizer) necessary to understand the results?
    \item[] Answer: \answerYes{} % Replace by \answerYes{}, \answerNo{}, or \answerNA{}.
    \item[] Justification: The paper specifies the training and test details in the experimental section and supplementary material.

\item {\bf Experiment statistical significance}
    \item[] Question: Does the paper report error bars suitably and correctly defined or other appropriate information about the statistical significance of the experiments?
    \item[] Answer: \answerNo{} % Replace by \answerYes{}, \answerNo{}, or \answerNA{}.
    \item[] Justification: The paper does not report error bars over multiple random seeds due to the high computational cost.

\item {\bf Experiments compute resources}
    \item[] Question: For each experiment, does the paper provide sufficient information on the computer resources (type of compute workers, memory, time of execution) needed to reproduce the experiments?
    \item[] Answer: \answerYes{} % Replace by \answerYes{}, \answerNo{}, or \answerNA{}.
    \item[] Justification: The paper reports the compute resources used for the experiments.
    
\item {\bf Code of ethics}
    \item[] Question: Does the research conducted in the paper conform, in every respect, with the NeurIPS Code of Ethics \url{https://neurips.cc/public/EthicsGuidelines}?
    \item[] Answer: \answerYes{} % Replace by \answerYes{}, \answerNo{}, or \answerNA{}.
    \item[] Justification: The research conforms to the NeurIPS Code of Ethics.

\item {\bf Broader impacts}
    \item[] Question: Does the paper discuss both potential positive societal impacts and negative societal impacts of the work performed?
    \item[] Answer: \answerYes{} % Replace by \answerYes{}, \answerNo{}, or \answerNA{}.
    \item[] Justification: The paper discusses broader impacts in the supplementary material.
    
\item {\bf Safeguards}
    \item[] Question: Does the paper describe safeguards that have been put in place for responsible release of data or models that have a high risk for misuse (e.g., pre-trained language models, image generators, or scraped datasets)?
    \item[] Answer: \answerNA{} % Replace by \answerYes{}, \answerNo{}, or \answerNA{}.
         \item[] Justification: The paper does not release high-risk generative models, scraped personal data, or assets with obvious dual-use risks.

\item {\bf Licenses for existing assets}
    \item[] Question: Are the creators or original owners of assets (e.g., code, data, models), used in the paper, properly credited and are the license and terms of use explicitly mentioned and properly respected?
    \item[] Answer: \answerYes{} % Replace by \answerYes{}, \answerNo{}, or \answerNA{}.
    \item[] Justification: The paper cites the existing assets used in this work and follows their terms of use.

\item {\bf New assets}
    \item[] Question: Are new assets introduced in the paper well documented and is the documentation provided alongside the assets?
    \item[] Answer: \answerYes{} % Replace by \answerYes{}, \answerNo{}, or \answerNA{}.
    \item[] Justification: The paper documents the new assets in the main paper and supplementary material.

\item {\bf Crowdsourcing and research with human subjects}
    \item[] Question: For crowdsourcing experiments and research with human subjects, does the paper include the full text of instructions given to participants and screenshots, if applicable, as well as details about compensation (if any)? 
    \item[] Answer: \answerNA{} % Replace by \answerYes{}, \answerNo{}, or \answerNA{}.
    \item[] Justification: The paper does not involve crowdsourcing experiments or human-subject studies.

\item {\bf Institutional review board (IRB) approvals or equivalent for research with human subjects}
    \item[] Question: Does the paper describe potential risks incurred by study participants, whether such risks were disclosed to the subjects, and whether Institutional Review Board (IRB) approvals (or an equivalent approval/review based on the requirements of your country or institution) were obtained?
    \item[] Answer: \answerNA{} % Replace by \answerYes{}, \answerNo{}, or \answerNA{}.
    \item[] Justification: The paper does not involve human-subject studies.

\item {\bf Declaration of LLM usage}
    \item[] Question: Does the paper describe the usage of LLMs if it is an important, original, or non-standard component of the core methods in this research? Note that if the LLM is used only for writing, editing, or formatting purposes and does \emph{not} impact the core methodology, scientific rigor, or originality of the research, declaration is not required.
    %this research? 
    \item[] Answer: \answerNA{} % Replace by \answerYes{}, \answerNo{}, or \answerNA{}.
    \item[] Justification: LLMs are not used as an important, original, or non-standard component of the core research.

\end{enumerate}
%%%%%%%%%%%%%%%%%%%%%%%%%%%%%%%%%%%%%%%%%%%%%%%%%%%%%%%%%%%%
\end{document}